\documentclass{article} % For LaTeX2e
\usepackage{iclr2027_conference,times}

\usepackage{amsmath,amsfonts,bm}

\def\eqref#1{equation~\ref{#1}}
\def\1{\bm{1}}

\DeclareMathAlphabet{\mathsfit}{\encodingdefault}{\sfdefault}{m}{sl}
\SetMathAlphabet{\mathsfit}{bold}{\encodingdefault}{\sfdefault}{bx}{n}

\usepackage{graphicx}
\usepackage{subcaption}
\usepackage{wrapfig}
\usepackage{hyperref}
\usepackage{url}
\usepackage[most]{tcolorbox}

\usepackage{float}
\usepackage{booktabs}
\usepackage{pifont}
\usepackage{tabularx}
\usepackage{multirow}

\newtcolorbox{promptbox}[2][]{
  enhanced,
  breakable,
  colback=gray!6,
  colframe=black!55,
  boxrule=0.5pt,
  arc=2mm,
  left=8pt,right=8pt,top=8pt,bottom=8pt,
  before upper={\setlength{\leftmargini}{0em}},
  title={#2},
  fonttitle=\bfseries\small,
  coltitle=black,
  colbacktitle=gray!15,
  attach boxed title to top left={xshift=0.8em,yshift*=-\tcboxedtitleheight/2},
  boxed title style={colframe=gray!15, colback=gray!15, boxrule=0pt, arc=1mm},
  #1
}

\title{Enabling Timely Guidance before Skill Retrieval: Retaining Helpful Warm Tips in Agent Context}

\author{Feng Liang$^1$, Yupeng Li$^2$, Runhao Zeng$^1$, Francis~C.~M.~Lau$^3$ \& Xiping Hu$^{1,4}$ \\
$^1$Shenzhen MSU-BIT University, $^2$Hong Kong Baptist University, \\
$^3$The University of Hong Kong, $^{4}$Beijing Institute of Technology
}

\newcommand{\method}{TipsWarm}

\iclrfinalcopy % Uncomment for camera-ready version, but NOT for submission.
\begin{document}

\maketitle
\fancyhead{}

\begin{abstract}
Reusable skills help LLM-based agents solve complex tasks, but the agent must receive guidance before it commits to an ineffective approach.
Existing skill mechanisms often expose only metadata and load full content on demand, leaving useful guidance unavailable until the agent decides to retrieve it.
General memory methods can incur substantial maintenance overhead, while keeping guidance in conversation context risks repeatedly exposing the agent to irrelevant or harmful advice.
We propose \method{}, a mechanism that complements existing skill mechanisms by maintaining a budgeted pool of skill-derived keypoints, or \textit{warm tips}, for selective injection into the context of every message turn.
By separating event-triggered LLM assessment from inexpensive per-turn screening, it makes transferable skill guidance readily available while controlling maintenance costs.
In three coding and iterative task-execution benchmarks, \method{} achieves the highest task success rate while remaining time-efficient, compared to recent skill and general memory baselines.
% It coordinates three maintenance operations with special considerations to both safeguard warm-tip quality to help transfer skill guidance to future tasks while not polluting the context, and remain efficient by co-designing the operation complexity and trigger timing.  
% On AppWorld, \method{} achieves 71.0\% task success within 40 turns, improving over the underlying skill mechanism by 4.9 percentage points with approximately 8\% higher mean completion time.
% On AlfWorld, it attains the highest mean final success rate among the evaluated methods while using approximately 42\% of Dynamic Cheatsheet's mean time to complete successful tasks.
\footnote{The source code and results are available at https://github.com/DistriAI/TipsWarm.}\label{footnote:code}
\end{abstract}

\begin{wrapfigure}{r}{0.5\textwidth}
\centering
\vspace{-1.5\baselineskip}
\includegraphics[width=\linewidth]{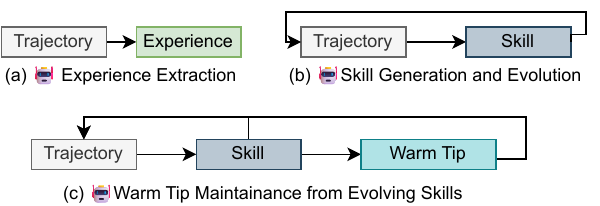}
\caption{Thumbnail comparison of different agent experience paradigms.}
\label{fig:thumbnail}
% \vspace{-0.5\baselineskip}
\end{wrapfigure}
\section{Introduction}\label{sec:intro}
Memory of experience is essential for LLM-based agents to accomplish complex tasks, 
such as coding~\citep{yang2024sweagent} and interactive decision making~\citep{yao2023react,shinn2023reflexion}.
% A skill structures reusable experience into actionable guidance that augments agents operating within a harness,
% helping them follow established guidelines or avoid known pitfalls~\citep{}.
A skill structures reusable experience into actionable guidance,
helping agents follow established procedures or avoid known pitfalls~\citep{li2026skillsbench}. 
Skills drafted manually in advance may miss keypoints or even provide misleading guidance for new tasks. 
% Considerable effort has been devoted to dynamically generating skills from agent experience, such as interaction trajectories~\citep{zhao2024expel,yang2026autoskill,ni2026trace2skill},
% with a hope that it is transferable and work well on similar tasks.
Considerable effort has been devoted to dynamically generating skills from agent experience, 
such as interaction trajectories~\citep{zhao2024expel,yang2026autoskill,ni2026trace2skill},
with a goal of producing guidance that transfers to similar tasks. 
As agents encounter new tasks, however, previously generated skills may prove flawed or become outdated. 
Recently, several studies have proposed methods for evolving skills throughout the task-solving process. 
For example, Memento-Skills~\citep{zhou2026mementoskills} updates and expands its skill library from new experience, 
and SkillRL~\citep{xia2026skillrl} recursively co-evolves a hierarchical skill bank with the agent policy.
These updates help skills adapt to new experience and improve their potential to transfer to future similar tasks.
% Basically, skills are long-term memory of agents' reusable experience, which reside in persistent storage.
% Skills thus serve as a form of long-term memory of agents' reusable experience and reside in persistent storage.
%  and are retrieved when needed.

A challenge of using skills is that useful experience can be buried in a large skill repository
and may not be retrieved in time to guide the agent's decisions.
This problem arises when the conversation context initially contains only skill metadata, such as names and descriptions,
and the agent retrieves full skill content only when it deems retrieval necessary.
% Before deciding to retrieve appropriate skill content, the agent can waste effort to explore the task space in a wrong direction,
% which prevents agents from solving tasks efficiently and achieving success effectively within budgets.
Before deciding to retrieve a relevant skill, the agent may waste effort on an ineffective approach,
reducing its chance of solving the task within available budgets of message turns.
Our hypothesis is that \textit{keeping relevant skill keypoints
readily available before full-skill retrieval helps agents 
make productive decisions earlier}.
This early guidance can help the agent avoid wasted effort and solve tasks more effectively.
For an analogy, imagine a programmer returning to experiment scripts written long ago, 
with the option specifications buried in a stack of documents.
The programmer may struggle to choose the right options even though the documentation is available for reference.
% Keeping a few lessons and warnings from similar experiments in mind can help the programmer choose suitable settings and recognize likely mistakes. 
The programmer may choose the wrong settings before realizing which documentation to consult. 
A short note beside the scripts, listing key constraints and common mistakes from previous experiments, 
can guide those choices immediately, while the full documentation remains available when needed.
% Without keeping the reminders of takeaways or pitfalls of different settings in mind, he/she can struggle and easily make mistakes in setting options for a new experiment, which can lead to failure or inefficiency. 
% can still struggle with recalling the options for a new experiment after some time, if he/she lacks previous example scripts in mind or to refer to at hand.
% Keeping example scripts in mind or at hand not only prevents using wrong option setting for experiments, but also increases task-solving efficiency without potentially additional reference.  
We describe helpful skill keypoints, such as the short note above, kept readily accessible in the conversation context as \textit{warm tips},
with a distinction from the \textit{cold skills} that must be dug up from memory when needed.
Retaining useful warm tips in the context can help guide the agent’s initial decisions before it recognizes the need to dig through the repository for a full skill.
Fig.~\ref{fig:thumbnail} compares different paradigms of experience memory in a thumbnail.

% Some existing in-context memory management mechanisms~\citep{suzgun2026dynamic,xu2025amem} can be considered as a variant of warm tips,
% Existing agent memory mechanisms have set foot in context management as alternatives to warm tips.
Existing agent memory mechanisms also manage reusable experience in the conversation context.
One general approach is to retrieve query-relevant information from memory and inject it into the context, 
independently of the skill mechanism.
For instance, A-MEM~\citep{xu2025amem} dynamically links and updates memory notes and retrieves query-relevant ones.
Dynamic Cheatsheet~\citep{suzgun2026dynamic} organizes reusable strategies and code snippets into a compact memory that conditions subsequent queries.
Managing experience separately from skills raises concerns about both effectiveness and efficiency.
On the one hand, a separate memory mechanism may overlook useful guidance already captured in skills,
including lessons and warnings that could prevent ineffective actions. 
On the other hand, maintaining a growing memory store and retrieving relevant information usually incur substantial overhead.
For example, according to our observations in preliminary experiments,
by updating, embedding, and retrieving information in almost every turn, 
A-MEM required twice as many LLM calls and took approximately 14-25$\times$ as much wall-clock time as the dynamic skill evolution baseline.
% Retaining a small set of warm tips from skills creates the opportunity to inherit their harness effects with much less overhead.
Drawing these tips from existing skills offers a way to reuse the guidance in existing skills with controlled additional overhead.

However, maintaining warm tips in context requires attention to both their quality and cost. 
Warm tips require more careful selection than stored skills because they can directly influence the agent's planning and actions in every turn.
Maintaining and injecting tips must also remain affordable.
While remaining compatible with the existing skill mechanism, warm tips should remain available in every turn for immediate guidance without requiring costly selection or updates each time.

We propose \textbf{\method{}}, an in-context warm-tip management mechanism with careful designs for effectiveness and efficiency.
% \method{} continuously maintains a budgeted warm-tip pool for quality and cost control through a loop procedure with three coherent operations.
\method{} continuously maintains a pool of candidate warm tips under a size budget and manages their quality through three operations.
First, when a skill is accessed, generated, or updated, the \textit{filtering} operation extracts keypoints
and admits those judged by an LLM to be potentially helpful for future similar tasks.
% The keypoints can hopely prevent the agent from planning and executing in a wrong direction when solving new similar tasks.
% The filtering operation schedules the warm tips within a budget-limited pool to limit the token cost,
% and the adimission and eviction criteria are based on an LLM judge's reasoning on the tips' transferability for similar tasks.
% Then, in every agent user message turn, the \textit{injection} operation selects high-utility warm tips from the pool to inject into the context.
Then, in every user message turn, the \textit{injection} operation selects warm tips from the pool based on their relevance and past usefulness and adds them to the context.
% This prevents polluting the context by known harmful tips, which we observe can mislead the agent and outweigh the benefits of helpful tips in preliminary experiments.
Because injected tips can influence subsequent actions, \method{} uses scope matching and historical utility estimates to reduce exposure to potentially misleading guidance.
% Finally, after task execution, the \textit{attribution} operation refines the contribution of injected tips by having an LLM to judge based on the task outcomes.
Finally, after task execution, the \textit{attribution} operation uses an LLM to assess how injected tips helped or misled the agent, based on its actions and task outcomes.
These assessments guide future tip selection during injection.
% The efficiency of \method{} is protected by a co-design of operation timing and strategy.
\method{} limits overhead by coordinating the timing and cost of these operations.
% The filtering and attribution operations, which use LLM judgments for complex reasoning are triggered only in low frequency,
% while the injection operation, triggered in every turn for prompt guidance, only relies on fast heuristic rules.
Filtering and attribution invoke an LLM only at skill events and task completion, respectively, while per-turn injection reuses their output through inexpensive heuristic rules.
% The pool can be persisted for reuse across conversation sessions.
% whose tips are filtered from persistent skills only at skill visit events
% and selectively injected in the conversation context in every user message turn. 
% \method{} is designed to be a complement of the existing skill mechanism
% and is integrated into the existing agent loop as a downstream component of the skill generation and evolution process.
\method{} complements the existing skill mechanism and integrates into the agent loop downstream of skill generation and evolution.
% Experiments on coding and interactive task-execution benchmarks demonstrate that
% \method{} can significantly enhance existing skill mechanism in solving similar tasks with high accuracy and efficiency.
Across three coding and interactive task-execution benchmarks, \method{} improves final task success over the existing skill mechanism. 
Compared to recent general memory methods, \method{} improves task success with better success-time trade-offs. 
% It also improves early-turn success on the coding tasks.  and efficiency of agents equipped with an existing skill mechanism when solving similar tasks.
Major contributions of this work are summarized as follows:
\begin{itemize}
    % \item We design \method{} to promptly guide agents to solve tasks in the right direction by injecting the context with the warm tips from skills.
% \method{} works as a downstream component of the existing skill mechanism to fully leverage the harness effects of existing skills.
\item We design \method{} to keep useful skill keypoints available in context, enabling them to guide agents before full skills are retrieved.
It complements existing skill mechanisms by reusing their guidance.
    % \item \method{} co-designs the strategy and timing to maintain a warm-tip pool for both effectiveness and efficiency.
% It uses LLM judges to extract potentially helpful warm tips and attribute their contributions in skill events that are not frequent,
% while keeping out harmful ones in the context through utility-aware fast screening in every turn.
\item \method{} coordinates the filtering, injection, and assessment of warm tips for both effectiveness and efficiency.
It uses LLM judgments only at skill events and task completion, and fast screening in every turn to limit agents' exposure to harmful tips. 
    % as a special kind of memory that are retrieved from persistent skills
    % and actively inject them in the conversation context in every user message turn.
    % \item We conduct extensive experiments on coding and interative desicion-making benchmarks and demonstrate that

\item 
% We conduct extensive experiments on coding and interactive task-execution benchmarks and show that
% \method{} achieves higher task success of agents using an existing skill mechanism
% and better success-time trade-off over general memory methods.
We show that selectively exposing existing skill guidance improves final task success across three benchmarks, 
and that \method{} enables agents with weaker LLMs to approach the performance of those with stronger ones and augments skills learned without warm tips.
\end{itemize}

% Warm tips in \method{} can be considered as a special kind of memory that are retrieved from persistent skills
% but maintained in the conversation context in every user message turn. 

\section{Related Work}
\paragraph{Agent experience.}
LLM agents commonly interleave reasoning with environment interaction~\citep{yao2023react}.  
Early approaches reuse experience at different levels of abstraction.  
Synapse stores and retrieves complete trajectories as in-context exemplars~\citep{zheng2024synapse}, 
whereas Reflexion converts task feedback into verbal reflections that are retained across trials~\citep{shinn2023reflexion}.  
Later work further abstracts experience into reusable procedural knowledge.  
For example, ExpeL~\citep{zhao2024expel} distills cross-task insights from successes and failures. 
Voyager~\citep{wang2024voyager} accumulates an embedding-indexed library of executable programs for lifelong embodied exploration.
Agent Workflow Memory~\citep{wang2025awm} induces recurring action routines for web agents.  
These methods established that compact procedures can transfer more effectively than replaying raw trajectories.  
They nevertheless generally retrieve experience or workflows in response to the current task.  
Our work studies a complementary failure mode where
an agent may plan incorrectly before it requests or successfully retrieves the relevant procedure.  
We therefore distill only decision-critical points from persistent skills 
and keep a budgeted subset continuously available in the working context.

% \paragraph{Automatic construction and evolution of agent skills.}
\paragraph{Skill generation and evolution.}
Recent systems treat natural-language instructions, code, and supporting artifacts as explicit skills 
that can be created and revised without updating the base model.  
One line of work focuses on skill synthesis.
Agent traces are distilled into reusable skills~\citep{alzubi2026evoskill}, consolidated across diverse traces~\citep{ni2026trace2skill}, or refined through verification~\citep{zhang2026coevoskills}.  
A second line studies continual skill evolution.  
These methods update skill libraries as new interactions arrive, either for an individual agent~\citep{yang2026autoskill,zhou2026mementoskills}, across users~\citep{ma2026skillclaw}, or across complementary forms of experience~\citep{jiang2026xskill}.  
A third line couples skills with model learning.
It uses evolving skills to guide policy optimization~\citep{xia2026skillrl} or distills skill guidance into model parameters~\citep{wang2026skillsd}.  
The above lines improve how skills are created, maintained, and learned,
whereas our work addresses a later stage in the skill lifecycle.  
\method{} is agnostic to the upstream skill generator and operates after skill creation by managing a small set of skill-derived warm tips that remains visible across turns.  
Rather than competing with their learning objectives, 
it can consequently augment skill libraries, whether they are static, automatically generated, or continually evolving.

\paragraph{Long-term memory and context management.}
One line of work equips agents with long-term memory and retrieves relevant records when needed.
Retrieval-augmented generation (RAG)~\citep{lewis2020rag} provides the canonical pattern of selecting records from non-parametric storage and concatenating them with the input.
Agent memory systems extend this pattern to episodic feedback~\citep{shinn2023reflexion}, learned action values~\citep{zhang2023rememberer}, and reusable workflows~\citep{wang2025awm}.  
A-MEM~\citep{xu2025amem} further organizes interactions as linked, evolving memory notes and retrieves relevant notes for each query.  
Although retrieval makes large memories scalable, its benefit depends on whether retrieval is initiated 
and whether the correct item ranks highly.  
A second line of work manages the information already placed in the context window.  
Models can underuse relevant evidence when it is poorly positioned in a long context~\citep{liu2024lost}, 
which motivates prompt-compression methods that preserve salient information under a token budget~\citep{jiang2023llmlingua}.  
Most closely related to our setting, Dynamic Cheatsheet~\citep{suzgun2026dynamic} continually organizes strategies, solution patterns, and code snippets 
into a compact memory that conditions subsequent queries.  

\method{} is a variant of both lines of work and is specially designed for skills.  
In contrast to A-MEM, which retrieves query-relevant interaction memories, \method{} proactively maintains skill-derived techniques in the active context without waiting for query-time retrieval.  
In contrast to Dynamic Cheatsheet, which learns a general memory directly from model inputs and outputs, \method{} extracts decision-critical points from explicit skill artifacts and schedules them within a fixed budget across agent conversation sessions.  
Its distinctive design is to inject this working set at every user turn so that critical techniques can influence planning before an agent decides to retrieve a full skill.  
Thus, persistent skills remain the comprehensive source of knowledge, 
while warm tips form the readily accessible layer between long-term memory and active reasoning.

\section{Background and Problem Definition}
\begin{figure}[!t]
\begin{center}
\includegraphics[width=\linewidth]{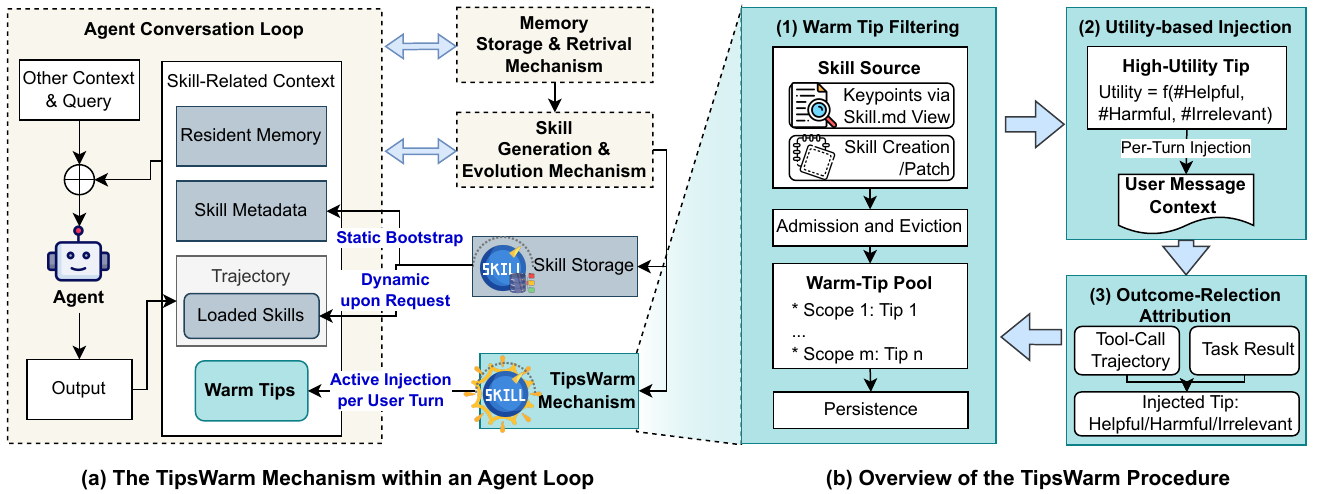}
\end{center}
\caption{Overview of a) \method{} in an agent conversation loop and b) \method{}'s operations}~\label{fig:overview}
\end{figure}

\paragraph{Background.} This study focuses on loop-based agent conversation~\citep{yao2025taubench},
which is the pervasive paradigm of general agent workflows as compared to the graph-based alternative~\citep{zhang2025aflow}.
An agent solves a task through a conversation session consisting of multiple turns.
Here, a turn refers to one cycle in which the agent processes a user message with the available context and produces a response.
We use ``loop'' and ``turn'' interchangeably.
A typical conversation loop is depicted in Fig.~\ref{fig:overview}~(a).
% In each turn, the agent inputs a user message with context and outputs a response,
% which can typically be a plan with reasoning or a tool-call request,
% including generating, loading or updating skills via skill tools.
% In each turn, the agent receives a user message with context and produces a response.
The response may contain reasoning, a plan, or a tool-call request,
including loading, generating or updating skills via skill tools.
The history of messages, agent responses, and tool results forms most of the conversation trajectory,
which then becomes part of the context for the next turn.

We next describe the context provided to the agent.
The conversation context includes both session-fixed and turn-specific information.
% A session-fixed prompt is fixed across turns and typically consists of system and user profile prompts,
% a full set of skill metadata,
% and other information that resides in the agent memory.
Session-fixed information typically includes system and user profile prompts.
Skill metadata and other memory content may also remain fixed if the agent does not refresh them during the session.
Turn-specific information is updated as the conversation progresses.
It mainly includes the conversation trajectory, which may contain skill content generated or loaded in previous turns.
In this study, we introduce warm tips as a new component of the turn-specific context.
Warm tips are keypoints extracted from skills and maintained in the context across turns.

\paragraph{Problem Definition.}
We formally define warm tips within an agent conversation session.
In turn $t$, we represent the context relevant to the skill and warm-tip mechanisms 
as a tuple $(\mathcal{P}, \mathcal{M}_t, \mathcal{T}_t, \mathcal{W}_t)$,
where $\mathcal{P}$ contains the system and user profile prompts, $\mathcal{M}_t$ contains the skill metadata,
$\mathcal{T}_t$ is the conversation trajectory before the current turn, 
and $\mathcal{W}_t$ is the set of warm tips injected in this turn.
Let a skill be $s = (m, \mathbb{C})$, with metadata $m$ 
and content represented as a set of skill points $\mathbb{C}=\{c_{1}, c_{2}, \ldots, c_{k}\}$.
Each skill point $c_i$ is a unit of guidance, such as an instruction, a constraint, or a warning about a pitfall.
The agent's skill repository at turn $t$ is denoted by $\mathbb{S}_t= \{s_1, s_2, \ldots, s_n\}$.
Both the number of skills and their contents may change over time.
The repository's skill metadata are $\mathcal{M}_t = \{m_i \mid (m_i, \mathbb{C}_i) \in \mathbb{S}_t\}$.
Skill metadata are typically loaded when the session starts and remain unchanged unless refreshed,
% The notation above assumes that the context contains the current repository metadata; if metadata are not refreshed, the agent instead receives the last loaded version.
while skill content can be generated or updated during the session.

% The conversation trajectory $\mathcal{T}_t$ can include previously loaded skill contents,
% that were generated or loaded in previous turns.
% These loaded skills,
% denoted as $\mathcal{S}_t \subseteq \mathbb{S}_{\leq t}$, that were accumulatively retrieved
% from previous sets of skills conditioned on the previous contexts and user messages.
The conversation trajectory $\mathcal{T}_t$ can include content from previously loaded skills.
Let $\mathcal{S}_t \subseteq \bigcup_{\tau<t}\mathbb{S}_{\tau}$ denote the skill versions loaded before turn $t$.
These versions were retrieved based on the contexts and user messages available in earlier turns.
% Let $\mathcal{U}_t$ be the user message at time $t$ and $f_{load}$ be the skill retrieval function.
Let $\mathcal{U}_t$ be the user message at turn $t$, and let $f_{load}$ represent the cumulative skill retrieval process.
% These loaded skills are therefore formulated as:
We summarize the process of loading skills below, where a subscript $<t$ denotes the corresponding history before turn $t$:
\begin{equation}
  \mathcal{S}_t = f_{load}(\mathbb{S}_{< t}; \mathcal{P}, \mathcal{M}_{< t}, \mathcal{T}_{< t}, \mathcal{W}_{< t}, \mathcal{U}_{< t}).
\end{equation}

In contrast, warm tips are selected skill points injected directly into the current context.
We denote them by $\mathcal{W}_t \subseteq \bigcup_{(m,\mathbb{C})\in\mathbb{S}_t}\mathbb{C}$, so each tip comes from a skill in the current repository.
% The continuous update of warm tips, $f_{warm}$, is conditioned on the current context, user message and precedent warm tips, which is formulated as:
Let $f_{warm}$ denote the warm-tip maintenance process.
It determines the tips for turn $t$ from the current skills, context, user message, and previously injected tips:
\begin{equation}
  \mathcal{W}_t = f_{warm}(\mathbb{S}_t; \mathcal{P}, \mathcal{M}_t, \mathcal{T}_t, \mathcal{W}_{t-1}, \mathcal{U}_t).
\end{equation}

Skills, warm tips, and memory are coherent but distinct in representation and meaning.
Loaded skills $\mathcal{S}_t$ provide full skill content, which may include complete procedures for solving tasks.
Warm tips $\mathcal{W}_t$ provide selected points from skills, such as useful techniques or warnings about pitfalls.
The skill points that exist in the loaded skills and warm tips may or may not overlap,
depending on the skill content and the warm-tip selection process.
% In addition, agents' general memory mechanism could develop variants for the context,
% which is independent of the warm-tip mechanism, which is a special long-term memory unique for skills.
General memory mechanisms may also supply information to the context.
The warm-tip mechanism specifically manages skill-derived guidance and can coexist with these mechanisms.

Given the context and user message, the LLM produces the agent response $\mathcal{O}_t$:
\begin{equation}
   \mathcal{O}_t = f_{LLM}(\mathcal{P}, \mathcal{M}_t, \mathcal{T}_t, \mathcal{W}_t, \mathcal{U}_t).
\end{equation}
The objective of this study is to design a warm-tip mechanism $f_{warm}$ 
that helps the agent solve new, similar tasks more efficiently and accurately.
It operates downstream of skill generation, updates, and retrieval, and can be used alongside general memory mechanisms.
The mechanism must select useful skill points while limiting the cost of maintaining and injecting them.

\section{Method}\label{sec:method}
\method{} consists of three major operations
% , including warm tip extraction, utility-aware injection, 
% and reflection-based attribution update,
as shown in Fig.~\ref{fig:overview}~(b).
1) The \textit{filtering} operation maintains a budget-limited pool of potentially helpful warm tips
with admission and eviction strategies for skill keypoints every time a skill is viewed, generated or updated.
2) The \textit{injection} operation quickly screens tips for relevance and utility, then adds the selected tips to the conversation context in every user message turn.
3) The \textit{attribution} operation assesses the contribution of injected tips using task outcomes and updates their utility records for future selection.
These operations connect skill-derived guidance, its use during a task, and feedback from that task.
% We present the detailed design of each component below.

% \paragraph{Design principles.} Effectiveness of warm tips and efficiency of maintaining them are both major concerns of \method{}.
\paragraph{Design principles.} \method{} must keep useful guidance available across turns (effectiveness) while limiting the cost of maintaining it (efficiency).
Refining skill quality can require substantial additional work. 
For example,
a typical approach is to iteratively generate
and evaluate candidate skills, retaining those that yield better
task performance~\citep{alzubi2026evoskill,zhang2026coevoskills}.
% a typical approach~\citep{zhang2026coevoskills} is to generate multiple skill variants
% and retain the one that performs best.
%  warm tips are filtered directly from existing skill mechanisms
% without extra refining.
1) In contrast, the first design principle of \method{} is to selectively reuse existing skills, 
which provides a source of guidance without a separate loop for generating and evaluating skill variants. 
\method{} therefore focuses its additional effort on selecting guidance for immediate exposure and updating its estimated usefulness from task feedback.
% whose benefits are twofold. 
% It not only inherits guidance produced by the upstream skill mechanism directly to support effectiveness, but also avoids the cost of generating and evaluating additional experience or skill variants for high efficiency.
% % The effectiveness is influenced by the quality of warm tips, or how instructive the tips are for solving future similar tasks.
% % \method{} maintains it by admitting and evicting tips beforehand
% %  beforehand criteria for admission and eviction 
% % and examining their contribution attribution afterward.
% % The effectiveness of warm tips depends on whether their guidance transfers to future similar tasks.
% The remaining effort is to prevent harmful tips from polluting the context through extra screening before injection in every turn and assessing the usefulness of tips after task outcomes. 
% \method{} assesses this potential when admitting and evicting tips, then revisits their usefulness after observing task outcomes.
% Warm tips are filtered directly from the existing skill mechanism 
% without extra refining. 
% This filtering operation not only
% takes advantage of the capability of various skill mechanisms for tips' quality,
% but is also much lighter-weight with manageable overhead than generating excessive redundant skill variants.
% \method{}'s efficiency is also protected by a careful design considering the complexity and timing of the operations.
2) A second design principle is to separate costly tip assessment from frequent tip injection for further runtime efficiency. 
% The filtering and attribution operations depend on LLM judgment to ensure semantic quality of the warm-tip pool,
% and are therefore triggered in low frequency only when visiting skills or finishing a task.
Filtering uses LLM judgment when skills are viewed, generated, or updated, while attribution uses it after task completion, 
both of which are triggered by relatively infrequent events. 
Injection runs in every turn using fast heuristic rules and the accumulated utility records, without an additional LLM judgment.
This allows immediate guidance to remain available throughout the conversation without repeating costly assessments in every turn. 
Below, we present the detailed design of \method{} for high effectiveness and efficiency.

% \textbf{
% scope is soft guidance.} soft gating.

\subsection{Warm Tip Filtering}
Warm tips in the conversation context should be kept concise and instructive.
% Full skill content is usually too large for the context window,
% and can contain irrelevant or even harmful information for new tasks.
Keeping full skill content in every turn can consume substantial context space and expose the agent to information that is irrelevant or even harmful for new tasks. 
% To avoid overwhelming and misleading skills in the context, we maintain a budget-limited pool of warm tips
% extracted from skills.
To limit context use and reduce misleading guidance, we maintain a budget-limited pool of warm tips extracted from skills.
% Based on our observations in preliminary experiments, the harm caused by misleading tips in the context may outweigh the benefits that helpful tips may bring.
Because injected guidance can influence successive decisions, a misleading tip may cause repeated wasted effort.
Therefore, we design this operation to guard against potentially harmful guidance while admitting tips likely to help on future similar tasks. 

Filtering maintains the warm-tip pool by combining efficient candidate extraction with LLM-based assessment of transferability.
First, it uses regular expressions (regex) to match skill sections that are heuristically considered important.
Examples include sections on pitfalls, best practices, and limitations, as well as explicitly tagged sections when the skill generator provides such tags. 
% Such regex-based filtering is efficient but can admit some harmful tips, which are dealt with in later operations.
Regex matching efficiently extracts candidate keypoints, but does not determine whether their advice is helpful or harmful.
Leaving out some helpful tips is less of a concern here, as the regular expression set can be extended to cover additional potentially helpful tips.
% The regex patterns can be extended to cover additional sections when useful guidance is missed.
% Then, an LLM-based admission and eviction strategy applies on the pool to control tips that may be beneficial for future similar tasks going into the pool
% and less helpful or even harmful ones being kicked out when the pool is full in budget.
Then, an LLM judges whether candidate tips are likely to help on unseen tasks of a similar kind.
The admission strategy adds promising tips to the pool, while the eviction strategy removes less useful or potentially harmful tips when the pool reaches its budget.
%  to further select 
% % from the filtered tips 
% tips that may be beneficial for future similar tasks and 
Warm tips are tagged with scope keywords for later injection screening.
The admission and eviction prompts are provided in Appendix~\ref{appx:prompt}.
The warm-tip pool is stored persistently for reuse across conversation sessions.
Conceptually, the filtering procedure can be represented as follows:
\begin{equation}
\text{Skills} \xrightarrow{\text{regex}} \text{Candidate Tips}
  \xrightarrow{\text{LLM judge with budget}} \text{Warm-Tip Pool}.
\end{equation}

% Admission and eviction of warm tips is a unified LLM-judged filtering operation that 
% is triggered only after a visit to skills.
% The filtering operation does not incur heavy overhead.
Filtering limits additional work by following the agent's existing skill operations.
% Its computational overhead is constrained because it is triggered only after a visit to skills,
% which happens occationally when the agent thinks it necessary to
% view or modify existing skills, or to generate new skills.
It runs only when the agent views, modifies, or generates a skill, so turns without these events require no filtering.
% The admission controls which tips go in the pool and the eviction controls which are kicked out when the pool is full in budget. 
% The space for the warm-tip context is contrained by the budget of warm-tip pool and further dynamic injection screening.
The pool budget limits the number of available tips, and injection screening to be introduced soon further restricts the subset added to the context. 
We cap the total number of tips in the pool (12 in our experiments) and the number contributed by each skill.
These limits control the amount of guidance retained and prevent a single skill from dominating the pool.

\subsection{Utility-aware Injection}
The injection operation selectively exposes the pool to the conversation context in every turn for immediate guidance.
% Though this slightly increases token cost per turn, the pool budget constrains the cost bound.
% In addition, as the overall token cost increases with conversation turns with trajectory growing in the Fibonacci-sequence style,
% reducing the number of turns by solving tasks in the right direction
% can easily offset the cost overhead per turn.
% In addition, as the conversation continues, later turns process a longer trajectory growing in the Fibonacci-sequence style.
% If warm tips help the agent solve a task in fewer turns, the resulting savings can offset the tokens added by the tips. 
The above admission of tips to the pool does not guarantee usefulness for the current task.
Injection therefore applies a second screening based on task relevance and historical utility to determine which tips in the pool enter the context.
Injected tips add tokens to each turn, so we constrain this overhead via the pool budget and injection quota. 
% to guide the agent in the right direction.

Because injection runs in every turn, relevance and utility are evaluated using inexpensive heuristics. 
We first match the scope keywords assigned during filtering against the task query to exclude tips from unrelated scopes.
% For each scope-relevant tip, we count its frequencies of being helpful, harmful, or irrelevant to past tasks,
% which are updated by the attribution operation
% to be introduced in the next subsection.
For each scope-relevant tip, we use counts of past tasks in which it was labeled helpful, harmful, or irrelevant,
which are updated by the attribution operation as described in the next subsection.
We derive the utility score of each tip simply by positively
weighting the helpful counts and negatively weighting the harmful and irrelevant counts.
We then select scope-relevant tips with high utility scores up to the injection quota and insert them immediately before the user message.
This reuses past assessments to guide current selection without asking an LLM to reassess the pool in every turn. 
Conceptually, the procedure for screening warm tips to be injected into the context can be represented as follows:
\begin{equation}
\text{Warm-Tip Pool} \xrightarrow{\text{scope match}} \text{Relevant Tips}
  \xrightarrow{\text{utility screening}} \text{Injected Tips}.
\end{equation}

\subsection{Outcome-Reflection Attribution}
% The correctness of tip utility relies on reasonable attribution of tip contribution
% using task execution outcome as feedbacks.
Useful utility scores require feedback on whether a tip helped or hindered task execution.
When several tips are injected together, the task outcome alone cannot distinguish their individual contributions.
A successful task does not imply that every tip helped, and a failed task does not imply that every tip was harmful.

% \method{} conducts LLM-reflection based on multi-dimensional outcomes together with tool-call events as clues to reason about tips' contribution.
Therefore, \method{} combines task outcomes with evidence from the agent's actions in an LLM-based assessment.
The judge considers success and, when available, efficiency and reliability measures alongside tool-call events that indicate whether the agent benefits from a tip.
Reprocessing the complete conversation trajectory for attribution can add non-negligible token cost. 
In comparison, a compact record of tool-call events retains key evidence of the agent's actions while omitting lengthy observations and raw tool outputs.
The judge uses this evidence to assess whether a tip plausibly helped or caused wasted effort.
In the attribution operation, each injected tip is counted as helpful, harmful, or irrelevant depending on LLM judgment,
whose prompt is provided in Appendix~\ref{appx:prompt},
and the other tips in the pool are counted as irrelevant.
Conceptually, the attribution procedure is:
\begin{equation}
(\text{Injected Tips, Tool-Call Events, Task Outcome})
  \xrightarrow{\text{LLM judge}} [\text{Helpful}|\text{Harmful}|\text{Irrelevant}].
\end{equation}
\begin{equation}
\text{Non-injected Pool Tips} \longrightarrow \text{Irrelevant}.
\end{equation}

\begin{figure}[!t]
\begin{center}
\includegraphics[width=\linewidth]{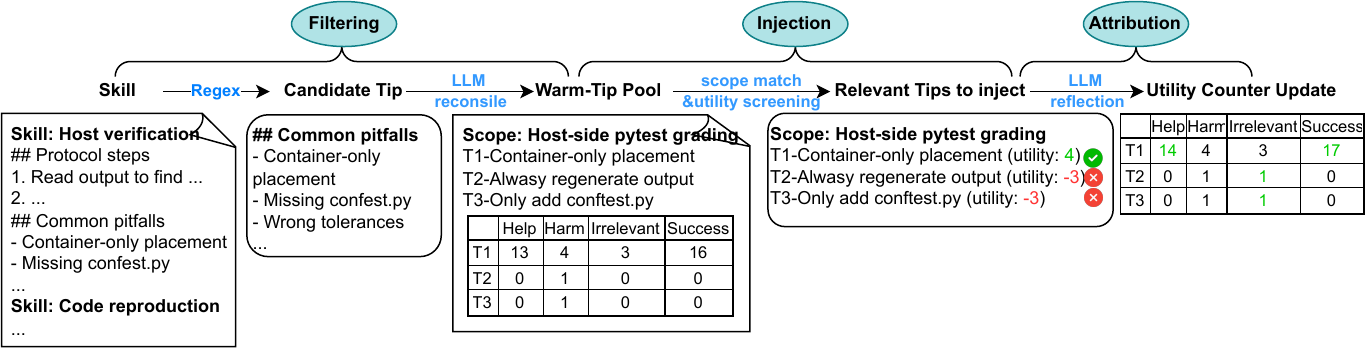}
\end{center}
\caption{An illustrative example of how warm tips are processed and represented across operations.}~\label{fig:example}
\end{figure}

For a concrete understanding of the above procedure, we illustrate the representation forms of warm tips across the operations in Fig.~\ref{fig:example}, 
using an example run on SkillsBench~\citep{li2026skillsbench}.

\section{Evaluation}
\subsection{Experiment Settings}\label{sec:expSetting}
\paragraph{Benchmarks and Splits.}
We use three multi-turn problem-solving benchmarks.
As evaluations should operate in the memory-equipped setting, 
all benchmarks use a warmup task split for initial skill generation and evolution and a test split of similar tasks for continuous memory evolution and performance evaluation.
\textbf{1) SkillsBench}~\citep{li2026skillsbench} is a comprehensive benchmark to evaluate agents' ability to leverage skills to solve coding tasks to pass test cases.
It contains 88 tasks across eight professional domains, solving which requires coding, debugging, and testing in development environments.
We randomly split the tasks in a 7:3 ratio stratified by domain, resulting in 65 warmup tasks and 23 test tasks.
\textbf{2) AlfWorld(-Text)}~\citep{shridhar2021alfworld} requires an agent to interact with a text-based game environment 
by taking recurring actions in space to complete household routines. 
% Exploring the space environment forms trajectories that can be distilled into skills. 
Considering the homogeneity in task actions and environment, we use only the first 50 tasks out of the total 3,553 tasks in AlfWorld's train set as warmup tasks, 
but all 134 tasks in its unseen set as test tasks for evaluation.
This setting can highlight the agent's ability to transfer skills distilled from a smaller set of warmup tasks to a large set of test tasks.
\textbf{3) AppWorld}~\citep{trivedi2024appworld} requires an agent to complete mobile app tasks by programming with various provided APIs.
It combines characteristics of the above two benchmarks, which involve coding and debugging to pass tests, as in SkillsBench
and exploration through interaction with an external environment, as in AlfWorld. 
All 90 tasks in the released train set are used as warmup tasks and all 168 tasks in the test-normal set are used as test tasks.

\paragraph{Metrics.} 
\textbf{1) Accuracy:}
% We use task-wise success rate within a turn budget $k$ (macro $Success@k$)
% test case-wise success rate (micro $Success@k$) to measure methods' accuracy of individual user turns, 
% accompanied by the Area under Curve of success rate from turn 1 to turn k ($AUC_{\leq k}$) to reflect the cumulative performance. 
We use task success rate within a turn budget $k$ ( $Success@k$) to measure the accuracy at individual turn budgets, 
accompanied by the Area under the Curve (AUC) of success rate from turn 1 to the maximum turn budget to reflect the cumulative performance. 
\textbf{2) Time Cost Efficiency:} We use the pair consisting of task success rate and task completion time ($Success@k, Time$) as a measure. 
Completion time is averaged over successful tasks only.
We also include results of token cost efficiency in Appendix~\ref{appx:tokenCost}, with no obviously consistent advantage across methods. 
% to within the maximum turn budget $k$ at the cost of the average number of tokens ($Success@k, \#Token$) and the average task completion time ($Success@k, Time$), respectively, to measure the cost efficiency. 
% The efficiency metrics for all benchmarks are the average number of tokens and turns to succeed. 
% Both the tokens and time only include those of successful tasks. 
% The cost of failed tasks is excluded from the efficiency metrics, but the number of successful tasks is indicated for appropriate interpretation.
% As AlfWorld tasks do have test cases and either succeeds or fails , it only uses the macro measures. 

% Among the benchmarks, SkillsBench is more challenging and often cannot be solved within the maximum allowed turns ($k=60$).
% Therefore, for SkillsBench with test cases, effectiveness metrics are both the task-wise pass rate before turn $k$ (macro \textit{Pass@k}) 
% and test case-wise pass rate (micro \textit{Pass@k}).
% While for AlfWorld and AppWorld where each task either succeeds or fails, 
% the effectiveness metric is the task success rate.

% , so the intrinsic efficiency difference between  of more effective methods that succeed more often before using up the maximum turns can be adjusted considerably higher from the literal metrics, especially when compared to less effective methods.
% The maximum allowed turn for all tasks is $k=60$.

\paragraph{Baselines.} We use the following baselines for skill or memory management, 
which are all integrated within the Hermes agent system~\citep{hermes_2026}, 
a ReAct-style~\citep{yao2023react} tool-using agent with an initial skill bundle and periodic skill generation and update.
\textbf{1) HermesSkill} represents Hermes' default skill mechanism that dynamically loads skills into the context 
upon explicit skill visiting operations. 
HermesSkill matches a direct comparison with \method{}, which retains its skill and memory mechanisms adding warm-tip management across turns.
Besides this, we compare with \textbf{2) A-MEM}~\citep{xu2025amem} and \textbf{3) Dynamic Cheatsheet (DC)}~\citep{suzgun2026dynamic},
which represent SOTA general memory baselines for long-term memory retrieval and context management, respectively. 
They upgrade the agent's default memory mechanism and can be considered as alternative counterparts to \method{}.

\paragraph{Running Details.}
We deploy Qwen3.6-27B~\citep{qwen36_27b_2026} as the agent's backbone LLM, 
with reasoning and tool-call abilities enabled and a maximum context length of 131,072.
% At most 12 tips can be retained in the pool, while at most 4 tips are allowed to be injected per turn.
A detailed list of major hyperparameters for \method{} is provided in Appendix~\ref{appx:hyperparam}.
For each experiment, after running once on the warmup tasks, 
the agent checkpoints its environment and proceeds to the test tasks.
Except for A-MEM, which densely updates its experience and takes significantly longer than other methods, 
all test experiments are run three times to report the average performance with standard deviation.
Considering the trajectory length required for each benchmark, the maximum allowed turns are 60 for SkillsBench, 50 for AlfWorld, and 40 for AppWorld.
All experiments are conducted on a server with a single NVIDIA A800 GPU with 80 GB of memory and 128-core CPUs of Intel Xeon Platinum 8468. 
The running costs of different methods on different benchmarks are listed in Appendix~\ref{appx:cost}.
\method{} uses a comparable amount of tokens but much less wall-clock time for warmup than the general memory methods (e.g., only about 1/22 of A-MEM and 1/3 of DC in AppWorld), as it avoids per-injection LLM judgment and additional experience management beyond the basic skill mechanism.

\subsection{Main Results}

\begin{wrapfigure}{r}{0.5\textwidth}
% \begin{figure}[!t]
\centering
\vspace{-3.2\baselineskip}
\begin{subfigure}[t]{0.54\linewidth}
    \centering
    \includegraphics[width=\linewidth]{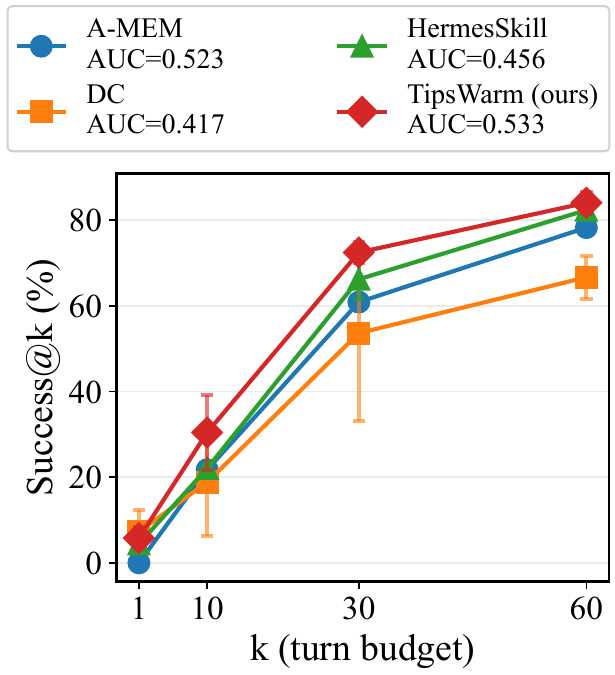}
    \caption{Accuracy}
    \label{fig:skillsbench-rate-macro}
\end{subfigure}\hfill
% \begin{subfigure}[t]{0.33\linewidth}
%     \centering
%     \includegraphics[width=\linewidth]{pic/appworld_micro_success_at_k}
%     \caption{Micro Accuracy}
%     \label{fig:app-rate-micro}
% \end{subfigure}\hfill
\begin{subfigure}[t]{0.46\linewidth}
    \centering
    \includegraphics[width=\linewidth]{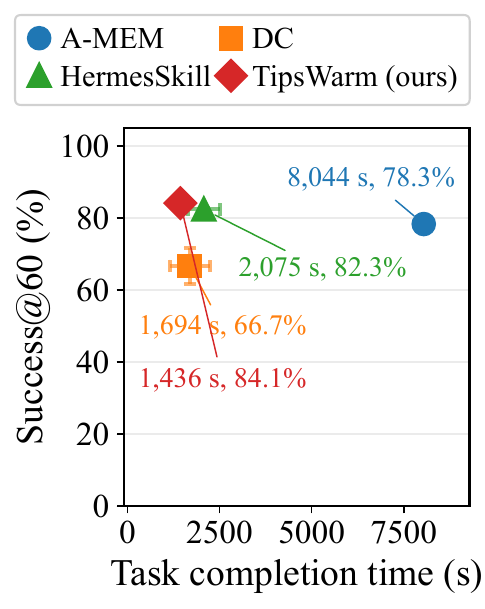}
    \caption{Time Cost Efficiency}
    \label{fig:skillsbench-efficiency-macro}
\end{subfigure}\hfill
\vspace{-0.5\baselineskip}
\caption{Performance on SkillsBench}
\label{fig:skillsbench}
\vspace{-0.5\baselineskip}
% \end{figure}
\end{wrapfigure}
\paragraph{SkillsBench.}
Fig.~\ref{fig:skillsbench} shows that \method{} improves both success rate and time cost efficiency on coding tasks requiring iterative debugging and testing.
Compared with HermesSkill, \method{} not only achieves a higher success
rate within 60 turns (84.1\% vs.\ 82.3\%) but also shows an advantage as early as
10 turns (30.4\% vs.\ 22.1\%), with AUC increasing from 0.456 to 0.533.
Meanwhile, \method{} reduces the task completion time by 30.8\%.
These suggest that \method{} helps agents avoid unproductive attempts by keeping tips about host-grading lessons, with examples shown in Appendix~\ref{appx:skillsbenchTip}, and that the added tip-management cost can be offset by more efficient task execution.
Against general memory baselines, \method{} 
% achieves 5.8 and 17.4 percentage points (pp) higher final success rate than A-MEM and DC, respectively, 
achieves a final success rate that is 5.8 and 17.4 percentage points (pp) higher than those of A-MEM and DC, respectively,
while also having the lowest completion time.
A-MEM approaches \method{}'s AUC but takes $5.6\times$ as much time on successful tasks.
% , which shows \method{}'s efficiency advantage by reusing the existing skill mechanism without separate experience management.
% DC's in-context cheatsheet falls short in final success.

\begin{wrapfigure}{r}{0.5\textwidth}
% \begin{figure}[!t]
\centering
\vspace{-1.2\baselineskip}
\begin{subfigure}[t]{0.51\linewidth}
    \centering
    \includegraphics[width=\linewidth]{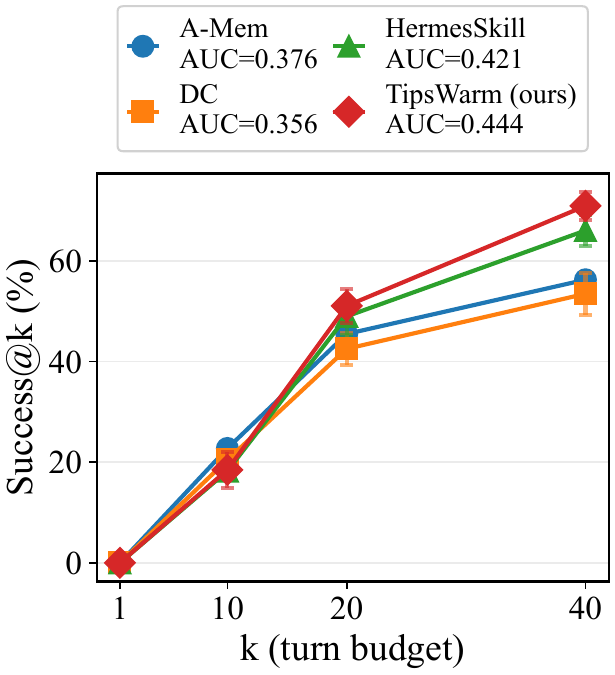}
    \caption{Accuracy}
    \label{fig:app-rate-macro}
\end{subfigure}\hfill
% \begin{subfigure}[t]{0.33\linewidth}
%     \centering
%     \includegraphics[width=\linewidth]{pic/appworld_micro_success_at_k}
%     \caption{Micro Accuracy}
%     \label{fig:app-rate-micro}
% \end{subfigure}\hfill
\begin{subfigure}[t]{0.49\linewidth}
    \centering
    \includegraphics[width=\linewidth]{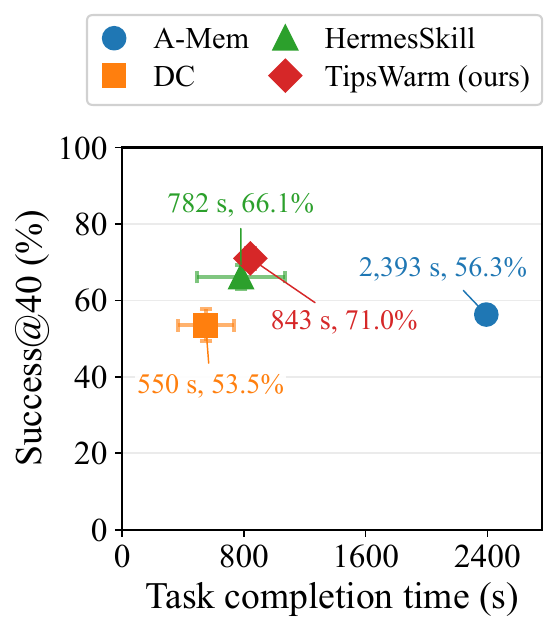}
    \caption{Time Cost Efficiency}
    \label{fig:app-efficiency-macro}
\end{subfigure}\hfill
\vspace{-0.5\baselineskip}
\caption{Performance on AppWorld}
\label{fig:appworld}
\vspace{-0.5\baselineskip}
% \end{figure}
\end{wrapfigure}
\paragraph{AppWorld.}
Fig.~\ref{fig:appworld} shows that \method{} extends the accuracy benefits to coding tasks involving interactions with application APIs. 
Within 40 turns, \method{} achieves the highest success rate of 71.0\%, exceeding HermesSkill and DC by 4.9 and 17.5 pp, respectively.
% Fig.~\ref{fig:appworld} shows that \method{} achieves the best accuracy within competitive cost for coding tasks interacting with APIs in AppWorld. 
% HermesSkill already beats both general memory methods on the task accuracy, and \method{} widens that gap.
% Within 40 turns, \method{} achieves the average task success rate of 71.0\%, 4.9 pp and 17.5 pp improvements over HermesSkill and DC, respectively. 
Unlike on SkillsBench, this improvement comes with a modest increase (7.9\%) in task completion time over HermesSkill. Nevertheless, \method{} remains $2.8\times$ as fast as A-MEM,
% Its task completion time is among the top-tier methods, alongside HermesSkill and DC, and is 2.9$\times$ as fast as A-MEM, 
which requires memory retrieval and synchronization every few turns, but finally carries lots of task-specific experience fragments into the context, as we observe.
% Interestingly, though all methods achieves similar test case accuracy, the task accuracy differs significantly. 
% AppWorld requires agents to coordinate APIs across apps and passing many individual tests need not imply satisfying all requirements of a task.
As shown in Appendix~\ref{appx:appworldTip}, \method{} can retain in the context
the reminders to check API parameters and resolve identifiers, providing concise, reusable guidance for avoiding errors across different tasks.

\begin{wrapfigure}{r}{0.5\textwidth}
\centering
% \vspace{-0.5\baselineskip}
\begin{subfigure}[t]{0.5\linewidth}
    \centering
    \includegraphics[width=\linewidth]{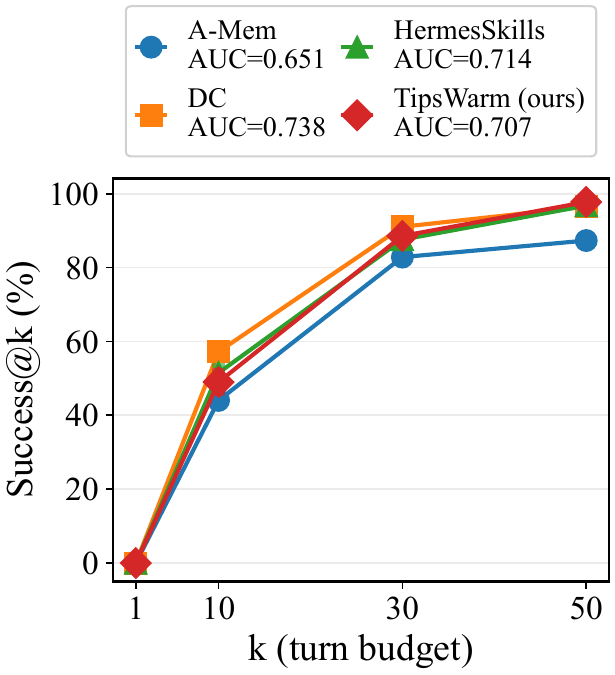}
    \caption{Accuracy}
    \label{fig:alf-rate-macro}
\end{subfigure}\hfill
\begin{subfigure}[t]{0.5\linewidth}
    \centering
    \includegraphics[width=\linewidth]{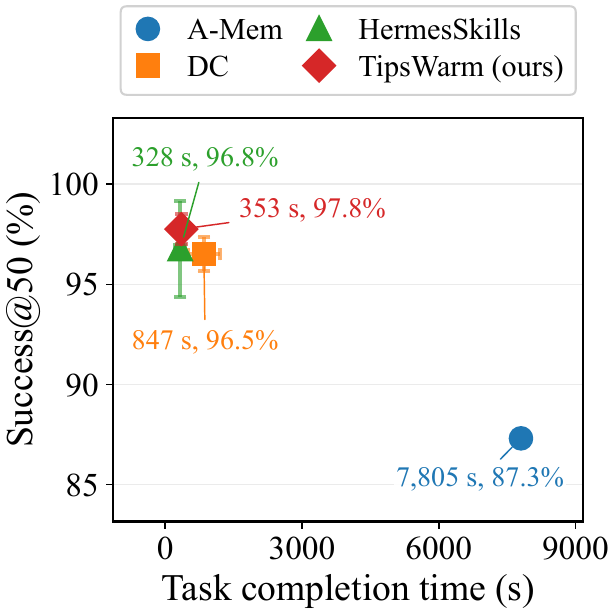}
    \caption{Time Cost Efficiency}
    \label{fig:alf-efficiency-macro}
\end{subfigure}
\vspace{-0.5\baselineskip}
\caption{Performance on AlfWorld.}
\label{fig:alfworld}
\vspace{-0.5\baselineskip}
\end{wrapfigure}
\paragraph{AlfWorld.}
Fig.~\ref{fig:alfworld} shows that \method{} can be significantly more time-efficient than the general memory baselines on tasks with repetitive action patterns while achieving a high final success rate. 
% AlfWorld tasks include recurring household action routines across different objects and locations,
% which makes procedural experience useful beyond the tasks from which it was acquired.
% The significant advantage over general memory baseline
% Adding further raises the success rate margin by about 1 pp while keeping competitive completion time.
% This is consistent with readily available skill guidance helping on some remaining tasks.
\method{} achieves 1.2 pp and 10.4 pp higher success rates than DC and A-MEM, respectively, with only about 2/5 and 1/22 as much task completion time.
% DC also retains reusable guidance in context~\citep{suzgun2026dynamic}, so this comparison highlights the cost of maintaining that guidance as well as its availability.
% Indeed, DC uses fewer turns on successful tasks in these runs, 
% yet takes longer in wall-clock time as it requires LLM-based assessment during per-turn injection, while \method{} avoids.
Though DC's AUC is slightly higher than that of \method{} (0.738 vs. 0.707), 
it comes at the cost of 2.4$\times$ longer test time and 58.2$\times$ longer warmup time (Appendix~\ref{appx:cost}).
For tasks with simple, repetitive interaction,
% DC's memory-management overhead highlights \method{}'s benefit of inexpensive per-turn screening.
DC's memory management may account for a larger share of runtime, making \method{}'s inexpensive per-turn screening particularly useful.
When HermesSkill already achieves a high success rate of 96.8\%, \method{} inherits its skill ability and further raises the rate by about 1 pp 
by adding tips about command format and anti-loop rules in the context (Appendix~\ref{appx:alfworldTip}).
% The high success rate of HermesSkill shows that the existing skill mechanism is already good enough for such environment, leaving little room for improvement.
% \method{} inherits such skill ability and further raises the success rate margin by about 1 pp 
% by adding warm tips about command format and anti-loop rules in the context (as shown in Appendix~\ref{appx:alfworldTip}).

\subsection{Ablation Study}

% \begin{figure}[t]
\begin{wrapfigure}{r}{0.5\textwidth}
\centering
\vspace{-3\baselineskip}
\begin{subfigure}[t]{0.49\linewidth}
% \begin{subfigure}[t]{1\linewidth}
    \centering
    \includegraphics[width=\linewidth]{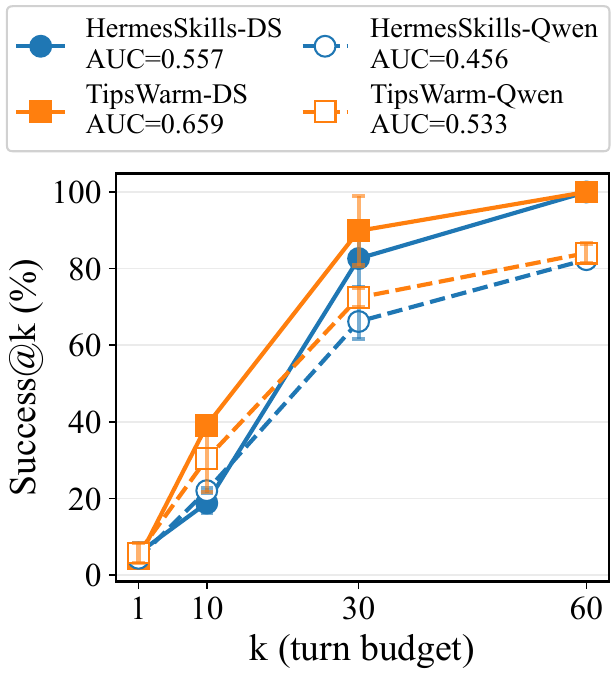}
    \caption{Accuracy}
    \label{fig:skillsbench-model-accuracy}
\end{subfigure}
% \begin{subfigure}[t]{0.49\textwidth}
\begin{subfigure}[t]{0.49\linewidth}
    \centering
    \includegraphics[width=\linewidth]{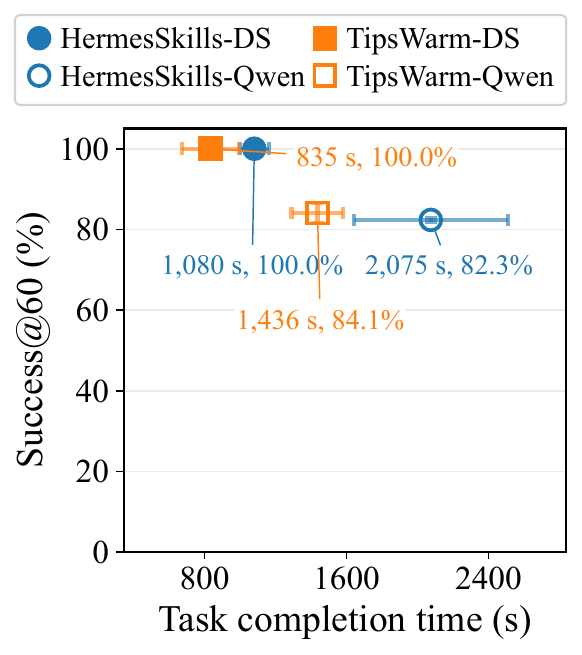}
    \caption{Time Cost Efficiency}
    \label{fig:skillsbench-model-cost}
\end{subfigure}
\vspace{-0.5\baselineskip}
\caption{Performance on SkillsBench using different backbone LLMs.}
\label{fig:skillsbench-model}
\vspace{-0.5\baselineskip}
% \end{figure}
\end{wrapfigure}
\paragraph{Backbone LLM.}
We change the backbone LLM to DeepSeek-V4.1-Flash (DS)~\citep{deepseekFlash26}, a recently released Mixture-of-Experts (MoE) model with higher reasoning and coding capability.
As Fig.~\ref{fig:skillsbench-model} shows, for SkillsBench tasks, the stronger DS can directly increase HermesSkill's final success rate to 100\% with an AUC of 0.557, corresponding to increases of 17.7 pp and 10.1 pp, respectively, over the results obtained with Qwen.
\method{} not only has the potential to shrink the capability gap between LLM models (\method{}-Qwen AUC 0.533 vs. HermesSkill-DS AUC 0.557), 
but can also further extend the capability margin of stronger models, increasing HermesSkill-DS's AUC by another 10.2 pp to 0.659. 
This indicates that \method{} can help agents with strong LLMs complete complex coding tasks in fewer turns, which also explains its time cost efficiency improvement over HermesSkill.

\begin{wraptable}{r}{0.48\textwidth}
\centering
\vspace{-1.2\baselineskip}
% \small
\caption{SkillsBench ablation results}
\label{tab:skillsbench-ablation}
\vspace{-0.5\baselineskip}
\setlength{\tabcolsep}{1pt}
\resizebox{\linewidth}{!}{%
\begin{tabular}{@{}lccccrr@{}}
\toprule
\multicolumn{1}{c}{Setting} & \shortstack{Warmup\\w/ Tips} & \shortstack{Scope\\Matching} & \shortstack{Utility\\Attribution} & \shortstack{Bigger\\Pool} & \multicolumn{1}{c}{\shortstack{Success@60 $\uparrow$\\(\%)}} & \multicolumn{1}{c}{AUC $\uparrow$} \\
\midrule
% NONE: run counts (success, AUC) = 3, 3
NONE &  &  &  &  & $82.3\pm0.5$ & $0.456\pm0.034$ \\
% NW: run counts (success, AUC) = 3, 3
NW &  & \ding{51} & \ding{51} &  & \underline{$85.5\pm2.5$} & {\boldmath $0.551\pm0.026$} \\
% NS: run counts (success, AUC) = 3, 3
NS & \ding{51} &  & \ding{51} &  & $84.1\pm2.5$ & $0.513\pm0.059$ \\
% NU: run counts (success, AUC) = 3, 3
NU & \ding{51} & \ding{51} &  &  & $82.6\pm0.0$ & $0.454\pm0.071$ \\
% BP: run counts (success, AUC) = 3, 3
BP & \ding{51} & \ding{51} & \ding{51} & \ding{51} & {\boldmath $88.4\pm2.5$} & $0.501\pm0.028$ \\
% ALL: run counts (success, AUC) = 3, 3
ALL & \ding{51} & \ding{51} & \ding{51} &  & $84.1\pm2.5$ & \underline{$0.533\pm0.023$} \\
\bottomrule
\end{tabular}%
}
\vspace{-0.5\baselineskip}
\end{wraptable}
\paragraph{Mechanism.}
Table~\ref{tab:skillsbench-ablation} shows \method{}'s accuracy on SkillsBench across various settings. 
\textbf{NW}: the agent inherits HermesSkill's warmed-up skill bundle.
\textbf{NS}: no scope matching during injection. 
\textbf{NU}: no utility attribution or screening. 
\textbf{BP}: a bigger pool containing at most 16 tips with up to eight tips injected per turn.
\textbf{NONE} and \textbf{ALL} are the default HermesSkill and \method{}.
As expected, scope matching and utility attribution help the agent to follow earlier guidance (NS vs. ALL and NU vs. ALL). 
A bigger pool can increase the final success rate but does not necessarily improve early-stage performance (BP vs. ALL).
Excitingly, \method{} can augment an existing skill bundle without requiring prior \method{}-style warmup in the first place (NONE vs. NW vs. ALL).

\section{Conclusion}
We presented \method{} to complements on-demand skill retrieval with warm tips selected for each conversation turn.
By separating event-triggered LLM assessment from fast per-turn screening, it makes skill guidance readily available while controlling context and maintenance costs.
Extensive evaluations on coding and iterative task-execution benchmarks show improved task success and success-time trade-offs over the underlying skill mechanism and general memory baselines.
% These findings support selectively retaining useful skill guidance as a practical complement to retrieving full skills.

% \begin{table}[t]
% \caption{Sample table title}
% \label{sample-table}
% \begin{center}
% \begin{tabular}{ll}
% \multicolumn{1}{c}{\bf PART}  &\multicolumn{1}{c}{\bf DESCRIPTION}
% \\ \hline \\
% Dendrite         &Input terminal \\
% Axon             &Output terminal \\
% Soma             &Cell body (contains cell nucleus) \\
% \end{tabular}
% \end{center}
% \end{table}

% \section{Default Notation}
% $a$ $\va$

\subsection*{AI use statement}
In this work, we used generative AI tools for implementing methods and providing feedback on experiments.
Additionally, we used generative AI tools for correcting grammar errors and improving readability. 
We have \textit{NOT} used generative AI tools to design or provide feedback on research  methodology,
assist with translation, clean and reformat datasets, support qualitative and thematic data analysis, or interpret results.
Generating synthetic datasets, helping develop theoretical models or conceptual frameworks, 
formulating mathematical claims, providing critical ingredients for proving mathematical claims, 
assisting in the writing of proofs, and proposing or refining hypotheses are \textit{NOT applicable} to this work.
We have reviewed all AI-assisted work. 
For example, LLM-generated code was verified and tested for correctness. 
We take responsibility for the final content of this work,
including text, claims or artifacts produced with the aid of generative AI.

\subsection*{Reproducibility statement}

% It is important that the work published in ICLR is reproducible. Authors are
% strongly encouraged to include a paragraph-long Reproducibility Statement at the
% end of the main text (before references) to discuss the efforts that have been
% made to ensure reproducibility. This paragraph should not itself describe
% details needed for reproducing the results, but rather reference the parts of
% the main paper, appendix, and supplemental materials that will help with
% reproducibility. For example, for novel models or algorithms, a link to an
% anonymous downloadable source code can be submitted as supplementary materials;
% for theoretical results, clear explanations of any assumptions and a complete
% proof of the claims can be included in the appendix; for any datasets used in
% the experiments, a complete description of the data processing steps can be
% provided in the supplementary materials. Each of the above are examples of
% things that can be referenced in the reproducibility statement.

We have taken steps to improve the reproducibility of our experiments. 
Section~\ref{sec:method} describes the method, 
and Section~\ref{sec:expSetting} describes benchmark splits, baselines, evaluation metrics, and execution setup.
Appendix~\ref{appx:prompt} provides the LLM assessment prompts, and Appendix~\ref{appx:settings} summarizes the main hyperparameters and computational costs.
The source code and experimental results are provided in the supplementary materials for anonymous review, which is indicated at the footnote in page~\pageref{footnote:code}.
The code documentation provides detailed instructions for (1) preparing the environment, (2) preparing benchmarks and data, (3) configuring baseline methods, (4) running experiments, and (5) locating the archived result summaries for the experiments reported in this paper.

\bibliography{iclr2027_conference}

@inproceedings{yao2025taubench,
  title={{$\tau$}-bench: A Benchmark for Tool-Agent-User Interaction in Real-World Domains},
  author={Yao, Shunyu and Shinn, Noah and Razavi, Pedram and Narasimhan, Karthik},
  booktitle={International Conference on Learning Representations},
  pages = {9965--10017},
  year={2025}
}

@inproceedings{zhang2025aflow,
 author = {Zhang, Jiayi and Xiang, Jinyu and Yu, Zhaoyang and Teng, Fengwei and Chen, XiongHui and Chen, Jiaqi and Zhuge, Mingchen and Cheng, Xin and Hong, Sirui and Wang, Jinlin and Zheng, Bingnan and Liu, Bang and Luo, Yuyu and Wu, Chenglin},
 booktitle = {International Conference on Learning Representations},
 pages = {34040--34077},
 title = {AFlow: Automating Agentic Workflow Generation},
 year = {2025}
}

@inproceedings{yao2023react,
  title={{ReAct}: Synergizing Reasoning and Acting in Language Models},
  author={Yao, Shunyu and Zhao, Jeffrey and Yu, Dian and Du, Nan and Shafran, Izhak and Narasimhan, Karthik and Cao, Yuan},
  booktitle={International Conference on Learning Representations},
  year={2023}
}

@inproceedings{shinn2023reflexion,
  title={Reflexion: Language agents with verbal reinforcement learning},
  author={Shinn, Noah and Cassano, Federico and Gopinath, Ashwin and Narasimhan, Karthik and Yao, Shunyu},
  journal={Advances in neural information processing systems},
  volume={36},
  pages={8634--8652},
  year={2023}
}

@inproceedings{yang2024sweagent,
  title={Swe-agent: Agent-computer interfaces enable automated software engineering},
  author={Yang, John and Jimenez, Carlos and Wettig, Alexander and Lieret, Kilian and Yao, Shunyu and Narasimhan, Karthik and Press, Ofir},
  journal={Advances in Neural Information Processing Systems},
  volume={37},
  pages={50528--50652},
  year={2024}
}

@inproceedings{zheng2024synapse,
  title={Synapse: Trajectory-as-exemplar prompting with memory for computer control},
  author={Zheng, Longtao and Wang, Rundong and Wang, Xinrun and An, Bo},
  booktitle={International Conference on Learning Representations},
  pages={19036--19066},
  year={2024}
}

@inproceedings{zhao2024expel,
  title={Expel: Llm agents are experiential learners},
  author={Zhao, Andrew and Huang, Daniel and Xu, Quentin and Lin, Matthieu and Liu, Yong-Jin and Huang, Gao},
  booktitle={Proceedings of the AAAI Conference on Artificial Intelligence},
  volume={38},
  number={17},
  pages={19632--19642},
  year={2024}
}

@article{wang2024voyager,
  title={Voyager: An open-ended embodied agent with large language models},
  author={Wang, Guanzhi and Xie, Yuqi and Jiang, Yunfan and Mandlekar, Ajay and Xiao, Chaowei and Zhu, Yuke and Fan, Linxi and Anandkumar, Anima},
  journal={arXiv preprint arXiv:2305.16291},
  year={2023}
}

@inproceedings{wang2025awm,
  title = 	 {Agent Workflow Memory},
  author =       {Wang, Zora Zhiruo and Mao, Jiayuan and Fried, Daniel and Neubig, Graham},
  booktitle = 	 {Proceedings of the 42nd International Conference on Machine Learning},
  pages = 	 {63897--63911},
  year = 	 {2025},
  volume = 	 {267},
  publisher =    {PMLR}
}

@inproceedings{lewis2020rag,
  title={Retrieval-augmented generation for knowledge-intensive nlp tasks},
  author={Lewis, Patrick and Perez, Ethan and Piktus, Aleksandra and Petroni, Fabio and Karpukhin, Vladimir and Goyal, Naman and K{\"u}ttler, Heinrich and Lewis, Mike and Yih, Wen-tau and Rockt{\"a}schel, Tim and others},
  journal={Advances in neural information processing systems},
  volume={33},
  pages={9459--9474},
  year={2020}
}

@inproceedings{zhang2023rememberer,
  title={Large language models are semi-parametric reinforcement learning agents},
  author={Zhang, Danyang and Chen, Lu and Zhang, Situo and Xu, Hongshen and Zhao, Zihan and Yu, Kai},
  journal={Advances in Neural Information Processing Systems},
  volume={36},
  pages={78227--78239},
  year={2023}
}

@article{liu2024lost,
  title={Lost in the middle: How language models use long contexts},
  author={Liu, Nelson F and Lin, Kevin and Hewitt, John and Paranjape, Ashwin and Bevilacqua, Michele and Petroni, Fabio and Liang, Percy},
  journal={Transactions of the association for computational linguistics},
  volume={12},
  pages={157--173},
  year={2024}
}

@inproceedings{jiang2023llmlingua,
  title={Llmlingua: Compressing prompts for accelerated inference of large language models},
  author={Jiang, Huiqiang and Wu, Qianhui and Lin, Chin-Yew and Yang, Yuqing and Qiu, Lili},
  booktitle={Proceedings of the 2023 conference on empirical methods in natural language processing},
  pages={13358--13376},
  year={2023}
}

@article{yang2026autoskill,
  title={Autoskill: Experience-driven lifelong learning via skill self-evolution},
  author={Yang, Yutao and Li, Junsong and Pan, Qianjun and Zhan, Bihao and Cai, Yuxuan and Du, Lin and Zhou, Jie and Chen, Kai and Chen, Qin and Li, Xin and others},
  journal={arXiv preprint arXiv:2603.01145},
  year={2026}
}

@article{alzubi2026evoskill,
  title={Evoskill: Automated skill discovery for multi-agent systems},
  author={Alzubi, Salaheddin and Provenzano, Noah and Bingham, Jaydon and Chen, Weiyuan and Vu, Tu},
  journal={arXiv preprint arXiv:2603.02766},
  year={2026}
}

@article{zhang2026coevoskills,
  title={Coevoskills: Self-evolving agent skills via co-evolutionary verification},
  author={Zhang, Hanrong and Fan, Shicheng and Zou, Henry Peng and Chen, Yankai and Wang, Zhenting and Zhou, Jiayu and Li, Chengze and Huang, Wei-Chieh and Yao, Yifei and Zheng, Kening and others},
  journal={arXiv preprint arXiv:2604.01687},
  year={2026}
}

@article{ni2026trace2skill,
  title={Trace2skill: Distill trajectory-local lessons into transferable agent skills},
  author={Ni, Jingwei and Liu, Yihao and Liu, Xinpeng and Sun, Yutao and Zhou, Mengyu and Cheng, Pengyu and Wang, Dexin and Zhao, Erchao and Jiang, Xiaoxi and Jiang, Guanjun},
  journal={arXiv preprint arXiv:2603.25158},
  year={2026}
}

@article{zhou2026mementoskills,
  title={Memento-skills: Let agents design agents},
  author={Zhou, Huichi and Guo, Siyuan and Liu, Anjie and Yu, Zhongwei and Gong, Ziqin and Zhao, Bowen and Chen, Zhixun and Zhang, Menglong and Chen, Yihang and Li, Jinsong and others},
  journal={arXiv preprint arXiv:2603.18743},
  year={2026}
}

@article{ma2026skillclaw,
  title={Skillclaw: Let skills evolve collectively with agentic evolver},
  author={Ma, Ziyu and Yang, Shidong and Ji, Yuxiang and Wang, Xucong and Wang, Yong and Hu, Yiming and Huang, Tongwen and Chu, Xiangxiang},
  journal={arXiv preprint arXiv:2604.08377},
  year={2026}
}

@article{jiang2026xskill,
  title={Xskill: Continual learning from experience and skills in multimodal agents},
  author={Jiang, Guanyu and Su, Zhaochen and Qu, Xiaoye and Fung, Yi R},
  journal={arXiv preprint arXiv:2603.12056},
  year={2026}
}

@article{xia2026skillrl,
  title={Skillrl: Evolving agents via recursive skill-augmented reinforcement learning},
  author={Xia, Peng and Chen, Jianwen and Wang, Hanyang and Liu, Jiaqi and Zeng, Kaide and Wang, Yu and Han, Siwei and Zhou, Yiyang and Zhao, Xujiang and Chen, Haifeng and others},
  journal={arXiv preprint arXiv:2602.08234},
  year={2026}
}

@article{wang2026skillsd,
  title={Skill-sd: Skill-conditioned self-distillation for multi-turn llm agents},
  author={Wang, Hao and Wang, Guozhi and Xiao, Han and Zhou, Yufeng and Pan, Yue and Wang, Jichao and Xu, Ke and Wen, Yafei and Ruan, Xiaohu and Chen, Xiaoxin and others},
  journal={arXiv preprint arXiv:2604.10674},
  year={2026}
}

@inproceedings{suzgun2026dynamic,
  title = "Dynamic Cheatsheet: Test-Time Learning with Adaptive Memory",
    author = "Suzgun, Mirac  and
      Yuksekgonul, Mert  and
      Bianchi, Federico  and
      Jurafsky, Dan  and
      Zou, James",
    booktitle = "Proceedings of the 19th Conference of the {E}uropean Chapter of the {A}ssociation for {C}omputational {L}inguistics (Volume 1: Long Papers)",
    month = mar,
    year = "2026",
    address = "Rabat, Morocco",
    publisher = "Association for Computational Linguistics",
    pages = "7080--7106"
}

@inproceedings{xu2025amem,
 author = {Xu, Wujiang and Liang, Zujie and Mei, Kai and Gao, Hang and Tan, Juntao and Zhang, Yongfeng},
 booktitle = {Advances in Neural Information Processing Systems},
 pages = {17577--17604},
 publisher = {Curran Associates, Inc.},
 title = {A-Mem: Agentic Memory for LLM Agents},
 volume = {38, Main Conference},
 year = {2025}
}

@article{li2026skillsbench,
  title={SkillsBench: Benchmarking how well agent skills work across diverse tasks},
  author={Li, Xiangyi and Liu, Yimin and Chen, Wenbo and You, Bingran and Di, Zonglin and He, Yifeng and Zheng, Shenghan and Choe, Kyoung Whan and Sun, Jiankai and Wang, Shuyi and others},
  journal={arXiv preprint arXiv:2602.12670},
  year={2026}
}

@inproceedings{
shridhar2021alfworld,
title={{\{}ALFW{\}}orld: Aligning Text and Embodied Environments for Interactive Learning},
author={Mohit Shridhar and Xingdi Yuan and Marc-Alexandre Cote and Yonatan Bisk and Adam Trischler and Matthew Hausknecht},
booktitle={International Conference on Learning Representations},
year={2021},
url={https://openreview.net/forum?id=0IOX0YcCdTn}
}

@inproceedings{trivedi2024appworld,
  title={Appworld: A controllable world of apps and people for benchmarking interactive coding agents},
  author={Trivedi, Harsh and Khot, Tushar and Hartmann, Mareike and Manku, Ruskin and Dong, Vinty and Li, Edward and Gupta, Shashank and Sabharwal, Ashish and Balasubramanian, Niranjan},
  booktitle={Proceedings of the 62nd Annual Meeting of the Association for Computational Linguistics (Volume 1: Long Papers)},
  pages={16022--16076},
  year={2024}
}

@url{qwen36_27b_2026,
  title = {Qwen3.6-27B},
  url   = {https://huggingface.co/Qwen/Qwen3.6-27B},
  year  = {2026}
}

@url{hermes_2026,
  title = {Nous Hermes Agent},
  url   = {https://github.com/nousresearch/hermes-agent},
  year  = {2026}
}

@misc{deepseekFlash26,
      title={DeepSeek-V4.1-Flash: Pushing the Limits of KV Cache Compression}, 
      author={DeepSeek-AI},
      year={2026},
      archivePrefix={arXiv},
      primaryClass={cs.CL}
}
\bibliographystyle{iclr2027_conference}

\newpage
\appendix

\section*{Appendix Contents}
\begingroup
\setlength{\parindent}{0pt}
\newcommand{\appendixentry}[2]{%
  \noindent\hyperref[#1]{\ref*{#1}\quad #2}\dotfill\pageref{#1}\par}
\appendixentry{appx:prompt}{Prompts for LLM-based Assessment}
\appendixentry{appx:settings}{Experiment Setting Details}
\hspace*{1em}\appendixentry{appx:hyperparam}{Hyperparameters}
\hspace*{1em}\appendixentry{appx:cost}{Running Cost}
\appendixentry{appx:tokenCost}{Results of Token-Cost Efficiency}
\appendixentry{appx:examples}{Examples of Warm Tips}
\hspace*{1em}\appendixentry{appx:skillsbenchTip}{SkillsBench Tips}
\hspace*{1em}\appendixentry{appx:appworldTip}{AppWorld Tips}
\hspace*{1em}\appendixentry{appx:alfworldTip}{AlfWorld Tips}
\endgroup
\bigskip

\section{Prompts for LLM-based Assessment}\label{appx:prompt}
This section lists three prompts used for LLM-based assessment in the filtering and attribution operations. 
The prompts are only used by a separate LLM to maintain the warm-tip pool but are not directly reflected in the conversation context for solving tasks. 

\paragraph{Admission judge.}
Every time there is a skill event, including viewing, generating, and updating skills, the filtering operation uses the following prompt to admit candidate warm tips. 

\begin{promptbox}[label={lst:hot-admit-sys}]{The Prompt for Admission Judge}
You gate admission of individual hot-skill keypoints to a broadcast pool.

Every admitted tip will be injected on later tasks, including held-out
tasks never seen while the pool was built. Admit only tips that remain
meaningful after stripping instance-specific tokens and would help on an
unseen task in the same benchmark/environment genre.

Return JSON only: \{"admit": ["id", ...]\} listing ids to admit from
the provided candidates. Omit ids for non-transferable tips. Prefer
admitting fewer strong tips over keeping marginal ones.

ADMIT tips that:
\begin{itemize}
\item[-] State evaluator-/environment-agnostic process rules (write graded
  artifacts where the harness reads them; verify on the graded surface;
  use admissible APIs/tools). That is not a default env closer call
\item[-] Describe reusable pitfalls without absolute paths, task IDs, or oracle
  shortcuts
\end{itemize}

REJECT tips that:
\begin{itemize}
\item[-] Tell the agent to run bundled solution/ scripts, read ground truth from
  tests, or skip builds using static expected constants
\item[-] Encode one episode's Docker/Lean/install recipe rather than a reusable rule
\item[-] Prescribe a termination ritual (always call done()/complete\_task,
  always return a short/minimal answer). Those are episode procedures
\item[-] Prescribe a process ritual (ALWAYS show\_api\_doc / check docs before
  every call) rather than a reusable constraint
\item[-] Would mislead on an unrelated later task
\end{itemize}
\end{promptbox}

\paragraph{Eviction judge.}
When the warm-tip pool lacks capacity for new candidates in the filtering operation, the following prompt determines which tips to evict from or retain in the pool.

\begin{promptbox}[label={lst:hot-evict-sys}]{The Prompt for Eviction Judge to Reconcile}
You curate a small hot-skill key-point pool used as short guardrail
reminders.

Hard constraint --- keep is the store, not a guaranteed inject: every
tip you KEEP may be prepended on later turns and on later tasks,
including held-out tasks that were never seen while the pool was
built. Inject later selects a matching subset. Optimize for
unseen-task transfer and broadcast-safety (still helpful or at least
harmless out of domain). Do NOT optimize for replaying the same
tasks that produced the tips, and do NOT maximize usefulness on the
present pool alone.

Return JSON only: \{"keep": ["id", ...]\} with at most \{keep\_n\}
ids drawn from the provided points. Do not invent ids. Prefer fewer
strong tips over filling the quota: return an empty or short keep
list when most candidates fail the bar.

KEEP tips that:
\begin{itemize}
\item[-] State a transferable pitfall or procedure that remains meaningful
  after stripping instance-specific tokens (paths, hostnames, accounts,
  filenames, task IDs, one-off product recipes) and would still apply
  to a new task instance in the same environment/genre
\item[-] Encode evaluator-/environment-agnostic process rules (e.g. write
  graded artifacts where the harness reads them; prefer admissible
  actions; use the provided API surface; verify before long runs)
  rather than a single task's setup recipe or a default env closer
\item[-] Add distinct failure-mode coverage --- prefer diversity of pitfalls
  across skills/topics when candidates are similar
\item[-] Use each point's natural-language scope (when present) as a soft
  hint. Prefer tips that are either broadly safe or clearly
  conditional; do not keep a narrow niche tip merely because it is
  useful on the current overflow
\item[-] When utility stats are present (n\_labeled \textgreater{} 0), prefer tips with
  more helpful than harmful attributions; when success rates are
  similar, prefer lower avg\_iterations\_when\_success / avg\_iterations
  (efficiency often separates tips when success is near-ceiling)
\end{itemize}

DROP or deprioritize tips that:
\begin{itemize}
\item[-] Are bound to absolute paths, hostnames, credentials, account names,
  phone numbers, one-off filenames, task IDs, or a single product/CLI/
  app workflow (unless the same sentence also states a general rule
  that survives removing those tokens)
\item[-] Recap one episode's entities or layout (which object was where,
  which inbox/account/file the last task used, numbered instance slots,
  exact names that will not recur). Identifier placeholders such as
  \textless{}path\textgreater{} or \textless{}id\textgreater{} do not make a recap transferable --- if the remaining
  claim is still a memory of one episode, drop it
\item[-] Teach oracle/leakage or harness-cheating shortcuts (reading hidden
  solvers/oracles, claiming pass without producing graded outputs,
  rewriting verifier paths to temporary locations as a default)
\item[-] Encode one task's I/O layout, auth bootstrap, or dependency install
  recipe rather than a reusable pitfall
\item[-] Act mainly as a cheatsheet for replaying an already-seen task
  (memorized steps, exact entities, or solutions that would not help a
  new unseen task in the same benchmark)
\item[-] Duplicate a stronger tip already being kept, or add a second tip
  from the same skill that covers the same failure mode
\item[-] Would mislead, waste steps, or pull work off the graded surface if
  shown on an unrelated later task
\item[-] Prescribe a default search order over instance-indexed slots
  (visit every shelf / drawer / file / message N) rather than a
  constraint that applies only after the current observation fails.
  Keep a legal-action rule if it can stand without the tour
\item[-] Prescribe a termination ritual (always call done()/complete\_task,
  always return a short/minimal answer). Those are episode procedures;
  the correct closer depends on the task. A NEVER/DO NOT about a bad
  closer is a constraint --- keep that
\item[-] Have clearly worse utility than alternatives (high harmful count,
  or much higher iteration cost for similar success)
\end{itemize}

Imperative intensity (NEVER/ALWAYS/MUST) is not evidence of quality ---
judge the abstract claim, not the wording.

About context: it describes the incoming extract / current task only
(often from the training task). Use it to understand what is
being admitted and to break ties. Do NOT treat "most relevant to
context", "will help if this task is repeated", or "likely needed
on similar seen tasks" as the primary keep criterion under
store-then-retrieve inject.

When utilities\_flat\_no\_helpful is true in the user payload, no tip
yet has a helpful attribution --- prefer dropping oracle/cheatsheet
candidates and episode-local recipes even if they are merely irrelevant.
\end{promptbox}

% ---------------------------------------------------------------------------
% 3. Outcome attribution  (build_outcome_attribution_messages)
% ---------------------------------------------------------------------------
\paragraph{Outcome-attribution judge.}
After task completion, the attribution operation uses the following prompt for LLM assessment to update tips' utility counters based on their contribution to the task outcome.

\begin{promptbox}[label={lst:hot-outcome-sys}]{The Prompt for Outcome Attribution Judge}
You attribute credit for task outcome to hot-skill tips that were
exposed during the episode (injected or opened via skill\_view).

The outcome is multi-dimensional --- do NOT use success alone. Consider
reward, iterations/steps (efficiency; lower is better when the task
succeeded), tests\_passed/tests\_total, and duration when present. On
benchmarks where success is near-ceiling, prefer judging whether a tip
helped or hurt efficiency and reliability.

episode\_log is a compressed action trace (tool names + short args,
env actions, or code API snippets). It does NOT include observations
or raw API dumps. Use it to see whether a tip was followed, ignored,
or contradicted. A tip whose advice never appears in the log is
irrelevant even on success. A tip that matches a long enumeration of
lookups or extra env steps is harmful when iterations are high.
Missing log fields are not evidence the tip helped.

When outcome.success is false OR outcome.claimed\_success\_mismatch is
true: you MUST NOT label any via=inject tip as helpful. Prefer harmful
for injected tips (they were broadcast into a failed episode).
History-only tips may be irrelevant.

When outcome.claimed\_success\_mismatch is true, the agent asserted
pass/completion in its final response but labeled evaluation failed ---
label injected tips that plausibly steered toward wrong-env verification,
oracle/replay shortcuts, or premature success claims as harmful.

Return JSON only: \{"labels": \{"\textless{}id\textgreater{}": "helpful"\textbar{}"harmful"\textbar{}
"irrelevant", ...\}\} for the given point ids. Do not invent ids.
\begin{itemize}
\item[-] helpful: tip is a transferable process or constraint that plausibly
  improved this episode and would still apply to a new task instance
\item[-] harmful: tip plausibly caused waste, wrong paths, or failure modes,
  or would mislead on a different instance (wrong object, path, account).
  On task failure, prefer harmful over irrelevant for injected tips that
  prescribe shortcuts, oracle replay, or environment-specific cheats
\item[-] irrelevant: tip did not meaningfully affect this episode, OR it is
  an episode-local recap (specific objects/locations/accounts/filenames)
  even if the episode succeeded --- those do not transfer
\end{itemize}

Judge the abstract claim, not wording. NEVER/ALWAYS/MUST is not
evidence of helpfulness. Prefer irrelevant over helpful when a tip
only restates one episode's entities.
If a tip plausibly caused extra iterations --- an enumeration tour,
or visiting slots the observation did not suggest --- label it
harmful even when the episode eventually succeeded.
\end{promptbox}

\section{Experiment Setting Details}\label{appx:settings}
\subsection{Hyperparameters}\label{appx:hyperparam}
Table~\ref{tab:hyperparam} summarizes the default hyperparameters
for running \method{}. 

\begin{table}[H]
\centering
\small
\caption{Default hyperparameters for running \method{}}
\label{tab:hyperparam}
\resizebox{\linewidth}{!}{%
\begin{tabular}{@{}llp{7.2cm}@{}}
\toprule
\textbf{Hyperparameter} & \textbf{Value} & \textbf{Description} \\
\midrule
\multicolumn{3}{@{}l}{\textit{\method{} mechanism}} \\
\addlinespace[2pt]
Pool size budget & 12 & Maximum number of candidate tips retained in the persistent pool. \\
Per-skill cap & 8 & Maximum tips admitted from any single skill, to avoid one skill dominating the pool. \\
Injection quota & 4 & Maximum tips inserted into the context of each user-message turn. \\
Scope match & lexical overlap & Exclude tips whose scope keywords do not overlap the task query. \\
Utility score & $+$helpful, $-$harmful/irrelevant & Rank remaining tips from attribution counts. Higher scores are injected first. \\
\midrule
\multicolumn{3}{@{}l}{\textit{Warmup and test}} \\
\addlinespace[2pt]
Backbone LLM & Qwen3.6-27B & For both agent and warm-tip judges, with reasoning and tool calling enabled. \\
Context length & 131{,}072 & Maximum context window of the deployed model. \\
Turn budget $k$ & 60 / 50 / 40 & Maximum agent turns on SkillsBench / AlfWorld / AppWorld. \\
% Warmup tasks & 63 / 50 / 90 & Tasks used to initialize skills and the warm-tip pool before testing. \\
% Test tasks & 24 / 134 / 168 & Held-out similar tasks for evaluation (SkillsBench / AlfWorld / AppWorld). \\
Protocol & warmup then checkpoint & The environment is saved after warmup and reused for the test split. \\
Repeats & 3 & Independent runs with mean~$\pm$~std reported (A-MEM is run once). \\
\bottomrule
\end{tabular}
}
\end{table}

\subsection{Running Cost}\label{appx:cost}
Table~\ref{tab:train-cost} summarizes the token, time, and space cost for different methods to initialize the skills or experience with the warmup tasks of various benchmarks. 
While the number of tokens varies across different methods and the storage space cost appears to be a minor practical concern, \method{} requires substantially less running time than the general memory methods. 

\begin{table}[H]
\centering
\small
\caption{Training cost of different baselines for completing warmup tasks of various benchmarks. 
The number of agent tokens is the per-task mean~$\pm$~std of
tokens required by the agent to solve tasks,
while the number of internal tokens is the number of tokens required by the baseline methods' internal operations. 
The time column reports the wall-clock duration required for the agent to finish the warmup tasks. Experience space is the persistent
tip/memory artifact in addition to the default skills written by each method.}
\label{tab:train-cost}
\resizebox{\linewidth}{!}{%
\begin{tabular}{llrrrrl}
\toprule
\textbf{Benchmark} & \textbf{Method} & \multicolumn{1}{c}{$N$} & \multicolumn{1}{c}{\shortstack{\textbf{\# Agent}\\\textbf{Tokens}}} & \multicolumn{1}{c}{\shortstack{\textbf{\# Internal}\\\textbf{Tokens}}} & \multicolumn{1}{c}{\textbf{Time}} & \textbf{Experience Space} \\
\midrule
\multirow{4}{*}{SkillsBench}
  & A-MEM  & 65 & $1.74{\pm}1.69$M & $3.1$K & 316.0 h
    & 	
2.2 MB: 65 notes, 115 KB; 2.1 MB sqlite3 data \\
  & DC     & 65 & $2.24{\pm}1.78$M & $3.2$K & 134.4 h
    & 84.7 KB: 5 cheatsheet items, 6.3 KB: 	
2 skills, 78.4 KB \\
  & HermesSkill & 65 & $2.53{\pm}1.65$M & --- & 113.1 h
    & 93.8 KB: 10 skills \\
  & \textbf{\method{} (ours)}    & 65 & $2.47{\pm}2.05$M & $2.8$K & 28.6 h
    & 105 KB: 	
2 skills, 103 KB; 2 tips, 1.7 KB\\
\midrule
\multirow{4}{*}{AppWorld}
  & A-MEM  & 90 & $3.79{\pm}3.54$M & $48.9$K & 	
261.5 h
    & 33.7 MB: 1482 notes, 2.5 MB; 31.2 MB sqlite3 data \\
  & DC     & 90 & $8.24{\pm}14.39$M & $4.2$K & 40.1 h
    & 20.2 KB: 4 cheatsheet items, 8.5 KB, 1 skill, 11.7 KB \\
  & HermesSkill & 90 & $3.74{\pm}3.33$M & --- & 47.3 h
    & 15.7 KB: 1 skill  \\
  & \textbf{\method{} (ours)}    & 90 & $3.91{\pm}3.05$M & $1.8$K & 12.2 h
    & 	
73.3 KB: 1 skill, 71.1 KB; 3 tips, 2.2 KB \\
\midrule
\multirow{4}{*}{AlfWorld}
  & A-MEM  & 50 & $0.67{\pm}0.76$M & $93.0$K & 76.9 h
    & 37.4 MB: 1763 notes, 2.5 MB; 34.9\,MB sqlite3 \\
  & DC     & 50 & $0.70{\pm}0.90$M & $13.1$K & 87.4 h
  & 16.2 KB: 7 cheatsheet items, 9.3 KB; 1 skill, 6.9 KB \\
  & HermesSkill & 50 & $0.54{\pm}0.51$M & --- & 1.5 h
    & 3.1 KB: 1 skill\\
  & \textbf{\method{} (ours)}    & 50 & $0.65{\pm}0.80$M & $1.4$K & 1.5 h
    & 6.5 KB: 1 skill, 3.6 KB; 6 tips, 3.0 KB \\
\bottomrule
\end{tabular}
}
\end{table}

\section{Results of Token-Cost Efficiency}\label{appx:tokenCost}
\begin{figure}[H]
\centering
% \vspace{-0.8\baselineskip}
\begin{subfigure}[t]{0.30\linewidth}
    \centering
    \includegraphics[width=\linewidth]{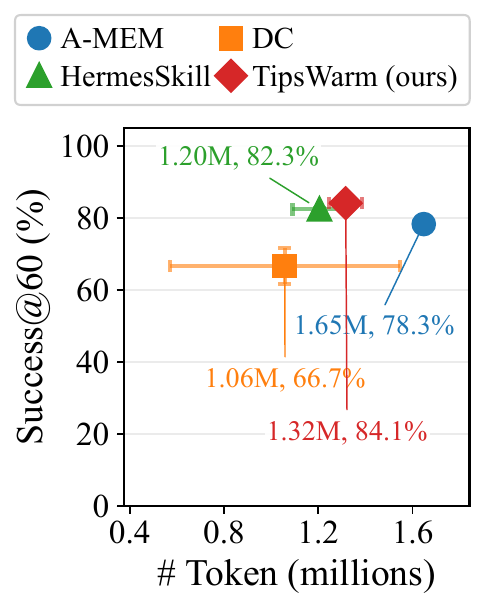}
    \caption{SkillsBench}
    \label{fig:skillsbench-rate-token}
\end{subfigure}
\begin{subfigure}[t]{0.34\linewidth}
    \centering
    \includegraphics[width=\linewidth]{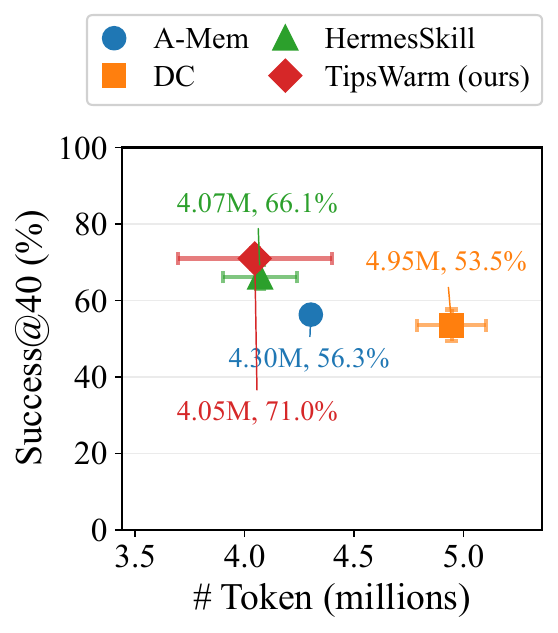}
    \caption{AppWorld}
    \label{fig:app-rate-token}
\end{subfigure}
\begin{subfigure}[t]{0.34\linewidth}
    \centering
    \includegraphics[width=\linewidth]{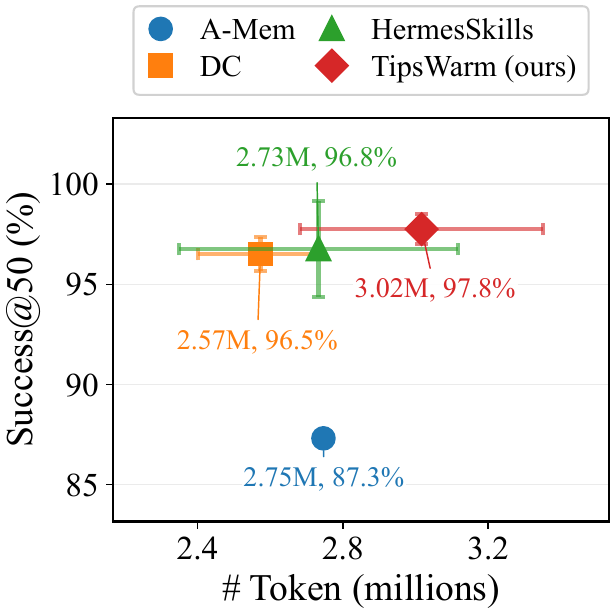}
    \caption{AlfWorld}
    \label{fig:alf-rate-token}
\end{subfigure}
\caption{Token-cost efficiency performance on various benchmarks}
\label{fig:token-cost}
\end{figure}

Fig.~\ref{fig:token-cost} shows the success rate within a turn budget and the corresponding number of tokens consumed by agents using different methods on various benchmarks.
Note that these agent tokens do not include tokens that may be required by methods' internal operations. 
No method consumes consistently fewer agent tokens than the others across all benchmarks.

\section{Examples of Warm Tips}\label{appx:examples}
We show examples of candidate warm tips remaining in the pool and their utility counters after all tasks are completed for different benchmarks.
The full content in the warm-tip pool also includes other descriptive metadata for tips, such as their scope, tag, and injection-related statistics, which we choose to omit here for brevity.
The quota of candidate tips that can enter the context is limited to four in the experiments. 

\subsection{SkillsBench Tips}\label{appx:skillsbenchTip}
Table~\ref{tab:skillsbench-warm-tips} shows a snapshot of the warm-tip pool
after a test run on the SkillsBench benchmark.
The pool contains five candidate tips belonging to a single skill.
Four of the five tips have a small positive utility (two helpful labels, no
harmful or irrelevant labels), and the remaining tip is unlabeled.
The final counters are sparse mainly for two reasons.
One is the scope screening of the injection operation, and the other is that
when any tip text is updated during reconciliation in the filtering operation,
the tip is considered a new tip and its counters are reset.
The helpful tips capture key requirements for host grading and artifact staging.
Without this guidance in advance, agents may incur substantial testing
and verification overhead.
\begin{table}[H]
\centering
\small
\caption{An example of candidate warm tips in the pool in a SkillsBench run}
\label{tab:skillsbench-warm-tips}
\begin{tabular}{@{}lrrr@{}}
\toprule
Tip & \#Helpful & \#Harmful & \#Irrelevant \\
\midrule
\multicolumn{4}{@{}l}{\textbf{Skill:} \texttt{skillsbench-host-grading}} \\
\addlinespace[2pt]
Graded files must be top-level task-dir entries (stager maps \texttt{/root}) & 2 & 0 & 0 \\
Mirror \texttt{/root/output/<f>} as top-level \texttt{output/<f>} & 2 & 0 & 0 \\
Do not reuse image-shipped \texttt{environment/} answers & 2 & 0 & 0 \\
Host pass needs \texttt{tests\_total$>$0} and \texttt{tests\_failed$=$0} & 0 & 0 & 0 \\
Stage \texttt{/root/results.json} as top-level \texttt{results.json} & 2 & 0 & 0 \\
\bottomrule
\end{tabular}
\end{table}

\subsection{AppWorld Tips}\label{appx:appworldTip}
Table~\ref{tab:appworld-warm-tips} shows the snapshot of the warm-tip pool
after a test run on the AppWorld benchmark.
The pool contains eight candidate tips belonging to two skills, also with sparse utility counters. 
The helpful tips focus mainly on API usage and identifier resolution. 

\begin{table}[H]
\centering
\small
\caption{An example of candidate warm tips in the pool in an AppWorld run}
\label{tab:appworld-warm-tips}
% \begin{tabularx}{lrrr}
% \begin{tabularx}{\linewidth}{@{}>{\raggedright\arraybackslash}Xrrr@{}}
\begin{tabular}{@{}lrrr@{}}
\toprule
Tip & \#Helpful & \#Harmful & \#Irrelevant \\
\midrule
\multicolumn{4}{@{}l}{\textbf{Skill:} \texttt{multi-app-api-workflow}} \\
\addlinespace[2pt]
Do not assume parameter names; check the API spec & 3 & 2 & 0 \\
Start from \texttt{show\_app\_descriptions()} & 0 & 0 & 0 \\
Never guess IDs; resolve them by search & 0 & 0 & 0 \\
Always call \texttt{show\_api\_doc} before use & 0 & 0 & 0 \\
\midrule
\multicolumn{4}{@{}l}{\textbf{Skill:} \texttt{api-sandbox-tasks}} \\
\addlinespace[2pt]
Tokens expire across code blocks (401 if reused) & 0 & 0 & 0 \\
Never invent IDs, tokens, or passwords & 3 & 2 & 0 \\
Paginate until the page is empty & 0 & 0 & 0 \\
Standard library only; no external clients & 0 & 0 & 0 \\
\bottomrule
% \end{tabularx}
\end{tabular}
\end{table}

\subsection{AlfWorld Tips}\label{appx:alfworldTip}
Table~\ref{tab:alfworld-warm-tips} shows the snapshot of the warm-tip pool
after a test run on the AlfWorld benchmark.
The pool contains eight candidate tips belonging to one skill.
These tips focus mainly on command format, action prerequisites, and avoiding repeated actions.
All tips have accumulated helpful labels, and none has been labeled harmful.
Each tip also has irrelevant labels, indicating that its usefulness varies across tasks.
These short reminders provide guidance applicable across household tasks with little additional context.

\begin{table}[H]
\centering
\caption{An example of candidate warm tips in the pool in an AlfWorld run}
\label{tab:alfworld-warm-tips}
\small
\begin{tabular}{@{}lrrr@{}}
\toprule
Tip & \#Helpful & \#Harmful & \#Irrelevant \\
\midrule
\multicolumn{4}{@{}l}{\textbf{Skill:} \texttt{alfworld-player}} \\
\addlinespace[2pt]
Use only commands from the admissible list & 84 & 0 & 32 \\
Reply with the bare command, with no quotes or prose & 84 & 0 & 32 \\
Find an object before trying to use it & 84 & 0 & 32 \\
Do not repeat a clean sequence that already finished & 33 & 0 & 25 \\
Stop after the goal is done; do not look or explain & 11 & 0 & 12 \\
Do not reopen a container just to check the object & 33 & 0 & 23 \\
Stop if the same command is repeated three times & 84 & 0 & 32 \\
Stop if open and close keep alternating & 84 & 0 & 32 \\
\bottomrule
\end{tabular}
\end{table}

\end{document}